\documentclass{article}
\usepackage{conference,times}

\usepackage{amsmath,amsfonts,bm}

\def\eqref#1{equation~\ref{#1}}
\def\1{\bm{1}}

\DeclareMathAlphabet{\mathsfit}{\encodingdefault}{\sfdefault}{m}{sl}
\SetMathAlphabet{\mathsfit}{bold}{\encodingdefault}{\sfdefault}{bx}{n}

\usepackage{hyperref}
\usepackage{url}
\usepackage{booktabs}
\usepackage{subcaption}

\usepackage{xcolor}
\usepackage{wrapfig}
\usepackage{xfrac}

\definecolor{Aqua}{HTML}{009ee3}
\definecolor{SkyBlue}{HTML}{59bdf7}
\definecolor{Fuchsia}{HTML}{e82e82}
\definecolor{Violet}{HTML}{983082}
\definecolor{Gray}{HTML}{706f6f}
\definecolor{Sunshine}{HTML}{ffbc29}
\definecolor{IndianRed}{HTML}{ec6469}
\definecolor{Caribbean}{HTML}{35cdb4}
\definecolor{Cornflower}{HTML}{4a4ad8}
\definecolor{Navy}{HTML}{0C122B}

\hypersetup{colorlinks=true, citecolor=Cornflower, linkcolor=red!50!black, urlcolor=green!50!black}

\newcommand{\legendline}[2]{\mbox{\tikz[baseline=-0.6ex]\draw[#1,#2,line width=1.5pt](0,0)--(0.35,0);}}

\usepackage[most]{tcolorbox}

\newcounter{finding}
\newcommand{\finding}[1]{
  \refstepcounter{finding}
  \begin{tcolorbox}[
    enhanced,
    breakable,
    colback  = Aqua!8,
    colframe = Cornflower,
    boxrule  = 0.8pt,
    arc      = 5pt,
    left=8pt, right=8pt, top=1pt, bottom=1pt,
  ]
    \textbf{Finding~\thefinding.}~#1
  \end{tcolorbox}
}

\newcommand{\boldheading}[1]{
\noindent\textbf{#1}}

\newcommand{\ourModel}{\textit{LoopICL}}
\newcommand{\TabArena}{\textit{TabArena}}
\newcommand{\Talent}{\textit{TALENT}}
\newcommand{\TabICLvTwo}{\textit{TabICLv2}}
\newcommand{\TabPFNvThree}{\textit{TabPFN-3}}
\newcommand{\TabPFNvTwo}{\textit{TabPFN v2}}

\newcommand{\ourModelNoRS}{($\alpha=1$, \legendline{Aqua}{solid})}
\newcommand{\ourModelDefualtRS}{($\alpha=\sfrac{1}{L}$, \legendline{Violet}{solid})}

\newcommand{\rqloopstack}{(RQ1) Can a single looped Transformer block match or exceed the predictive performance of the respective fixed-depth TFM?}
\newcommand{\rqrefine}{(RQ2) How does \ourModel{} refine its representations across recurrent steps?}
\newcommand{\rqperf} {(RQ3) How does \ourModel{}'s performance compare to substantially larger, fixed-depth frontier TFMs?
}

\title{LoopICL: Looping a single transformer block to solve tabular tasks}

\author{
Amir Rezaei Balef$^{1,2,3}$\quad Katharina Eggensperger$^{1,2}$ \\[0.4em]
$^1$TU Dortmund University, Dortmund, Germany \\
$^2$Lamarr Institute for Machine Learning and Artificial Intelligence, Dortmund, Germany \\
$^3$University of Tübingen, Tübingen, Germany \\
\texttt{amir.balef@tu-dortmund.de}
}

\iclrfinalcopy

\begin{document}

\maketitle

\begin{abstract}
Tabular foundation models using in-context learning have recently surpassed gradient-boosted trees on predictive tabular tasks.
However, recent mechanistic insights suggest that parameters in these models are largely redundant. We introduce \ourModel{}, a looped transformer whose core design decouples parameter count from computational depth. \ourModel{} consists of a single block, processing data through two coupled streams: a cell stream capturing per-cell feature representations and a row stream capturing in-context example representations, jointly refined through within-column and cross-column attention.
During pre-training, we vary loop counts,
allowing the block to be unrolled for a varying number of iterations at test-time and use a learned exit-gate to automatically exit. In its standard setting, \ourModel{} performs competitively with \TabICLvTwo{} on \TabArena{} and \Talent{} at the same computational cost (FLOPs), while using nearly $90\%$ fewer parameters. Furthermore, its recurrent design enables users to also trade off inference cost and performance, providing a resource-aware TFM.
\end{abstract}

\section{Introduction}

Tabular Foundation Models (TFMs) such as \TabICLvTwo{}~\citep{qu2026tabiclv2}, \TabPFNvThree{}~\citep{grinsztajn2026tabpfn}, and TabFM~\citep{tabfm2026} have surpassed gradient-boosted trees in predictive performance across a wide range of benchmark tasks \citep{liu2025talent, erickson2025tabarena, landsgesell2026scoringbench}. This highlights the potential of In-Context Learning (ICL) for supervised tabular tasks: given labeled training examples as context, these Transformer-based models can directly predict labels for new examples, without requiring task-specific training~\citep{hollmann-iclr23a}.

Current frontier TFMs rely on ICL and use a transformer-based model pre-trained on synthetic data generated from a prior by minimizing the loss on a given ICL task. As a result, these prior-data-fitted networks~\citep{muller-iclr22a} amortize predictive inference and generalize to real-world tasks out of the box. Specifically, they rely on deep stacks of independently parameterized transformer layers, resulting in fixed inference costs regardless of task complexity or computational budget. Recent progress has come from better input encoding and re-designing the prior (e.g., \TabICLvTwo{};~\cite{qu2026tabiclv2}), but is mostly based on continuously scaling up the parameter count (e.g., \TabPFNvThree{} is $7\times$ larger than \TabPFNvTwo{};~\cite{grinsztajn2026tabpfn}).

Mechanistic analyses suggest that these models are depth-wise redundant and mainly work through iterative refinement, e.g., increasing separability between classes in the latent space~\citep{balef2026is}. This finding is further strengthened by extensive compressibility via task-aware block pruning~\citep{koshil2026tacticl}.

Here, we challenge the scaling-up trend and use the insight that a transformer's computational depth is not necessarily tied to the number of unique weights and layers. So-called looped transformers replace a conventional stack of independently parameterized layers with repeated applications of the same transformer block \citep{dehghani-iclr19a}, thus decoupling parameter count from depth. This parameter sharing enables the model to increase its effective computational depth through iterative refinement while keeping the parameter count fixed, with the number of iterations adjustable at inference time to trade computational cost for predictive performance \citep{geiping2026scaling}.

This mechanism seems well suited for tabular tasks: Once tokenized, a supervised tabular task can be approached through iterative refinement of the latent representation, followed by a lightweight decoder to label query points. This differentiates tabular ICL from knowledge-intensive NLP tasks, which rely on retrieving learned facts and thus benefit from larger parameter counts~\citep{morrismuch}, and is consistent with the observation that existing TFMs exhibit different latent-space dynamics and redundancy compared to LLMs~\citep{balef2026is}.

\textbf{Contributions.} We propose \ourModel{}, to the best of our knowledge, the first recurrent Transformer architecture demonstrated at scale for tabular ICL, combining parameter sharing with iterative computation. At its core is a \textit{single transformer block} that contains a cell stream capturing per-cell feature representations and a row stream capturing in-context example representations, which is repeatedly applied to refine internal representations. A novel distributional conditioning on the input tokens stabilizes training and further improves predictive performance. On standard benchmarks, our model pushed the Pareto front of the accuracy–parameter trade-off, matching or exceeding the performance of 9-times larger TFMs (e.g., \TabICLvTwo{}).

\begin{figure}[tb]
\centering
\includegraphics[width=0.40\linewidth]{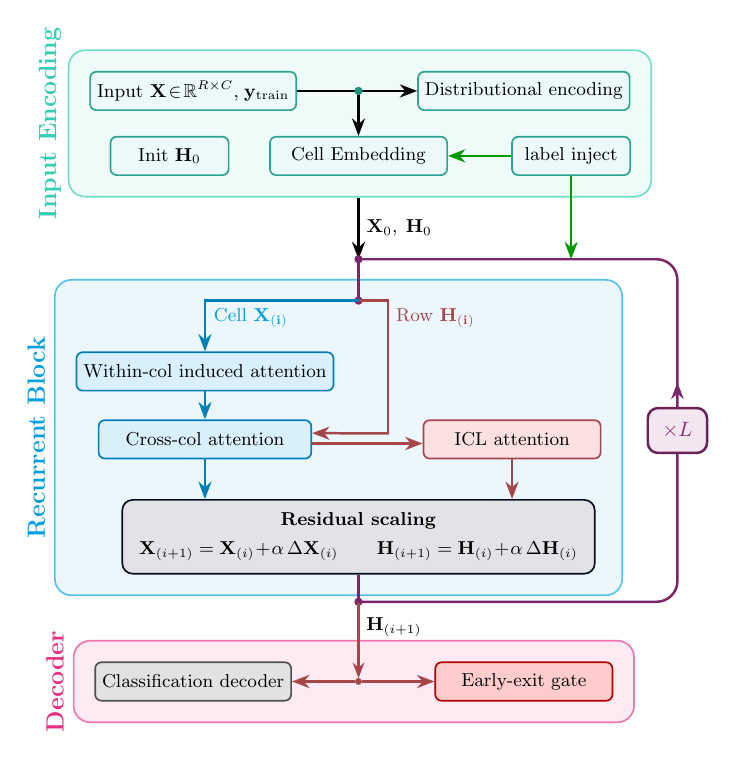}
\hfill
\includegraphics[width=0.55\linewidth]{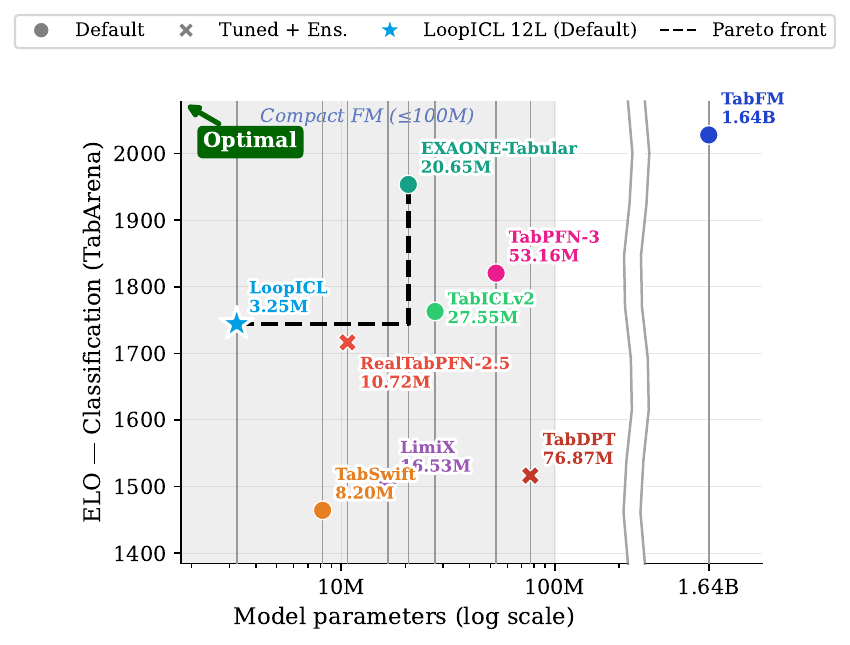}
\caption{Left: \ourModel{} architecture overview. The blue boxes operate on the \textcolor{Aqua!80!black}{\textbf{cell stream} (per-cell representations)}; the red box operates on the \textcolor{IndianRed!70!black}{\textbf{row stream} (per-row ICL representations)}. Right: \ourModel{} is highly parameter-efficient by reusing the same parameters across recurrent steps.}
\label{fig:architecture_params}
\end{figure}

To understand the behavior and capabilities of \ourModel{}, we conduct a detailed empirical analysis addressing the following research questions.

\textbf{\rqloopstack{}} We first establish recurrent models as a novel and competitive alternative for solving predictive tabular tasks.

\textbf{\rqrefine{}} We show that explicit residual scaling enables smooth refinement and extrapolation to high loop counts, and implicitly learned scaling enables the model to exit early at the minimum loops needed.

\textbf{\rqperf{}} Finally, we position our contribution in the context of the cost-efficiency of frontier TFMs.

\section{Related Work and Background}
\label{gen_inst}

\ourModel{} connects three lines of prior work. First, as a \emph{tabular foundation model} (TFM), it builds on transformer architectures for ICL on tabular data. Second, its recurrent design relates to \emph{looped} and weight-tied Transformers, which replace independently parameterized layers with repeated applications of the same block. Third, we study the behavior and scaling of recurrent computation, where varying the number of recurrent iterations provides an inference-time compute budget, connecting \ourModel{} to \emph{adaptive computation} and compute-efficient inference.

\boldheading{Related TFM Architectures.} TFMs are pretrained Transformer-based models that leverage ICL for predictive tabular tasks. Given a set of labeled examples as context, the models predict labels for new examples directly from these examples, without requiring task-specific parameter updates \citep{muller-iclr22a}. \textit{TabPFN} \citep{hollmann-iclr23a} pioneered this paradigm using a vanilla encoder-only architecture. Building on this design, \textit{TabSwift} \citep{liu2026tabswift} added gated attention for built-in support for Early Exit (EE). In 2025, \cite{hollmann-nature25a} introduced \textit{TabPFNv2}, which use 2D-attention transformers \citep{kossen2021self} applying attention along the row and column axes.

\textit{TabICL} \citep{qu2025tabicl} uses dedicated compression stage, followed by an encoder-only transformer to scale to larger contexts. \textit{TabICL v2} \citep{qu2026tabiclv2} further introduces query-aware scalable softmax attention for better generalization to larger datasets, an architecture subsequently adopted and improved by \TabPFNvThree{} \citep{grinsztajn2026tabpfn}.

\textit{EXAONE} \citep{eo2026exaonetabular10} introduces feature-summary and item-summary tokens, achieving state-of-the-art performance.

\boldheading{Recurrent tabular models.} The idea of iterative refinement in tabular prediction dates back to classical ensemble methods such as gradient boosting \citep{friedman-as01a}

successively fitting weak learners to the residuals of previous rounds. Their strong performance on tabular benchmarks \citep{grinsztajn-neurips22a,mcelfresh-neurips23a} provides evidence that this mechanism is well suited for tabular tasks. Also deep learning methods have explored iterative computation through recurrent and state-space architectures. \textit{Mambular}~\citep{thielmann2024mambular} and the SSM-based TFM of \citet{koch2025state} use state-space models for tabular prediction, while \citet{padayachy2026tab} applies the \textit{Tiny Recursive Model}~\citep{jolicoeur2025less} to insurance pricing and \citet{komisarczyk2026tydra} proposes a hybrid attention--state-space architecture. These works show that iterative computation can serve as the main principle for tabular tasks, further reinforced by the observation that even fixed-depth TFMs develop an inductive bias toward iterative refinement~\citep{balef2026is}.

\boldheading{Recurrent Transformers.} Recurrent Transformers repeatedly apply shared blocks to intermediate representations, increasing computational depth without increasing the number of parameters \citep{dehghani-iclr19a}.
\citet{saunshi2025reasoning} show that looped models can achieve strong reasoning performance with substantially fewer parameters and exhibit an inductive bias toward reasoning rather than memorization. Similarly, \citet{yang2024looped} demonstrate that weight sharing across iterations enables to efficiently emulate iterative learning algorithms.

Fixed-Point Reasoners interpret the process as convergence toward a stable representation and use convergence-based halting to allocate computation adaptively \citep{movahedi2026fixed}. Similar ideas have also been explored in visual generation~\citep{goyal2026elt} and audio processing~\citep{kaloga2026test}.
In the tabular domain, \citet{balef2026is} provide a small-scale proof of concept for using recurrence to iteratively refine latent representations. Collectively, these studies establish recurrence as a mechanism for increasing effective computation while maintaining parameter efficiency, however, it remains unclear how to scale computation to achieve competitive performance with large contexts.

\boldheading{Understanding recurrent Transformers.}
Several works suggest that recurrent computation progressively refines and stabilizes representations across iterations \citep{movahedi2026fixed, blayney2026mechanistic}, while the training-time distribution of recurrent steps shapes representation quality and generalization to unseen depths \citep{geiping2026scaling}. Mechanistically, \citet{blayney2026mechanistic} find the same recurrent block can perform different stages of computation across iterations. This suggests that each iteration can progressively carry out a different part of the overall computation, even though the same block is reused. In the tabular domain, \citet{balef2026is} similarly observe transitions in representations across TFM layers, suggesting distinct stages of computation. However, how representations evolve under recurrent computation in tabular ICL remains unexplored. We therefore investigate the dynamics of \ourModel{} across recurrent steps.

\boldheading{Adaptive computation.} Recurrence has traditionally been used to model sequential dependencies in recurrent neural networks \citep{elman1990finding}, but it has also been recognized as a mechanism for adaptive computation \citep{graves2016adaptive} and for introducing recurrent depth into Transformers \citep{dehghani-iclr19a}. Adaptive computation can be monitored via representation stability \citep{movahedi2026fixed} or prediction convergence \citep{geiping2026scaling}. Dedicated prediction heads can determine when to halt computation \citep{zhu2025scaling}. More recently, \citet{jeddi2026loopformer} showed that adaptive-computation strategies can be fragile when applied to looped Transformers and proposed budget-conditioned reasoning, where the user specifies the computational budget at inference time. \citet{Kuken2025EarlyStopping} propose an early-exit strategy for TFMs that use an entropy threshold as a proxy metric to exit the forward pass.

More recently, \citet{liu2026tabswift} proposed a per-query early-exit inference method, in which each query is processed independently, and the exit decision is made on a per-query basis. Their approach attaches lightweight prediction and exit heads to a subset of Transformer layers, enabling intermediate predictions and input-dependent stopping decisions.

\section{Architecture}
\label{architecture}
Figure~\ref{fig:architecture_params} (Left) shows an overview of the architecture; below, we highlight the key architectural innovations and components going through input encoding, residual scaling, and output decoding, while Appendix~\ref{app:architecture} provides full architectural details.

\boldheading{\textcolor{Caribbean}{Distributional Input Encoding.}} Raw features are standardized per column and encoded via cyclic feature grouping into $E$-dimensional cell embeddings~\citep{qu2026tabiclv2}. We inject class labels via an orthogonally initialized embedding matrix \citep{grinsztajn2026tabpfn} and re-inject them at each recurrent step. Before the recurrent loop, we additionally introduce \textit{distributional encoding}, which uses three complementary distribution-aware signals: \emph{marginal histograms}, \emph{discriminative histograms}, and \emph{Fourier quantile encodings}. These histogram-based signals capture marginal and class-discriminative feature distributions, while Fourier quantile encodings capture relative feature rankings and provide information about potential out-of-distribution test samples.

\boldheading{\textcolor{Aqua}{Recurrent Block.}} \ourModel{} processes data through two asymmetrically coupled streams: a \emph{cell stream} of individual cell representations, $\mathbb{R}^{R\times C\times E}$, and a compact \emph{row stream}, $\mathbb{R}^{R\times D}$, with $E=128$ and $D=4E$. This separation preserves fine-grained feature interactions while enabling efficient row-wise ICL, as relying solely on compressed row representations can lead to information loss~\citep{eo2026exaonetabular10}.
At each recurrent iteration, the cell stream first models local feature interactions and is then read out into the row stream. The row representations are refined through in-context attention, while the updated cell stream is retained and reused in the next iteration.

More specifically, each iteration applies a single shared axial block: within-column self-attention operates independently on each column, while cross-column self-attention with RoPE positional encodings operates across each row~\citep{su2024roformer}. The cross-column self-attention updates the cell stream, followed by a CLS-based readout into the row stream, which is then updated by causally masked ICL attention. All sub-blocks use sandwich normalization (pre- and post-RMSNorm)~\citep{ding2021cogview}.

\boldheading{\textcolor{Gray}{Residual Scaling.}} Since \ourModel{} applies the same block for $L \in \mathbb{N}^+$ iterations, residual scaling can improve training stability by controlling the accumulation of updates~\citep{wang2026residual,movahedi2026fixed}. After each iteration, given the block output $\tilde{\mathbf{x}}$, it constrains the update with alpha:
\begin{equation}
\mathbf{x} \leftarrow \mathbf{x} + \alpha (\tilde{\mathbf{x}}-\mathbf{x}),
\end{equation}
where $\alpha \in (0,1]$ is a configurable scaling factor, comparable to the step size in an optimization procedure; $\alpha=1$ means no scaling, and smaller values reduce the magnitude of each update, preventing the residual stream from diverging as $L$ increases. A fixed $\alpha$ may not generalize across highly varying $L$ values and encourages the model to greedily front-load computation into early iterations and actively suppress updates in later iterations. Alternatively, a depth-normalized scaling scheme that sets $\alpha$ as a function of a pre-defined loop depth $L$ encourages the model to spread computation uniformly across the given budget. To observe the impact of this in practice, we empirically study three variants ranging from no decay to linear depth normalization: $\alpha \in \{1, \sfrac{1}{\sqrt{L}}, \sfrac{1}{L}\}$.

\boldheading{\textcolor{Violet}{Decoder.}} Following \citet{grinsztajn2026tabpfn}, we use an attention-based retrieval decoder that is target-permutation equivariant \citep{arbel2026equitabpfn}. The normalized row-stream representations of test examples serve as queries and those of training examples serve as keys, with attention scores accumulated separately per class via one-hot label weighting \citep{koshil2025context}. We then average the resulting per-class attention masses across heads to produce class probabilities. Additionally, we introduce an early-exit gate that predicts whether further iterations are likely to improve performance. A compact MLP uses only training-set representations, combining the current state, its rate of change, the iteration index, and dataset size to produce an exit probability $\lambda \in (0,100)$.\footnote{Query-based early exit, as widely used in LLMs, may no directly transfer to TFMs. Unlike autoregressive LLMs, TFMs process all context elements in a single forward pass. Even if confident predictions are available for some query points, looping continues till all query points can be predicted. Consequently, we base our early-exit decision on the embeddings of the training samples and exit once iterative refinement converges.} Further details are provided in Appendix~\ref{app:earlyexit_gate}.

At test time, we use an ensemble of $N{=}8$ members, each receiving a differently transformed view of the data (permuted feature order, shuffled class labels, varied preprocessing); predictions are averaged across members. The early-exit gate operates independently per member, so each member may exit at a different loop step. Full ensemble details are provided in Appendix~\ref{app:inference}.

\subsection{Pretraining}
\label{sec:pretraining}

We pretrain \ourModel{} in two stages on synthetic tasks generated from the \TabICLvTwo{} graph-based SCM prior \citep{qu2026tabiclv2}. Here, we summarize the key settings, full details are in Appendix~\ref{app:pretraining}.

\begin{wrapfigure}{r}{0.3\linewidth}

\centering
\includegraphics[width=\linewidth]{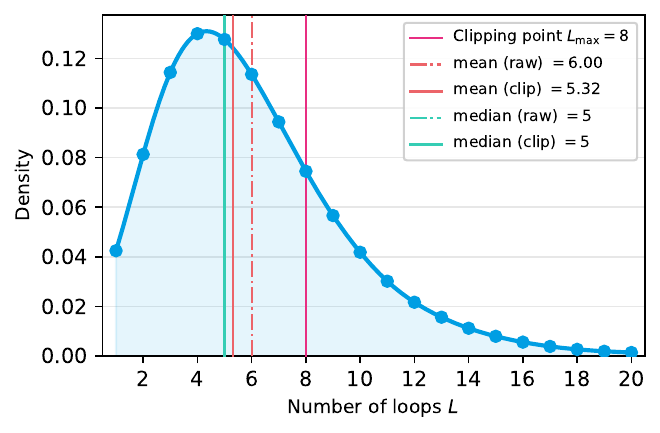}
\caption{The number of recurrent iterations $L$ is sampled from a log-normal Poisson distribution during pretraining.}
\label{fig:icl_loops_dist}
\vspace{-2em}

\end{wrapfigure}

\boldheading{Recurrent loop schedule.} The number of recurrent iterations for each training step is sampled from a log-normal Poisson distribution~\citep{geiping2026scaling} to improve depth generalization and performance retention~\citep{schwarzschild2021can, yang2026stabilizing}. We sample $\gamma \sim \mathrm{LogNormal}(\mu,0.5^2)$ and $L\sim 1 + \mathrm{Poisson}(\gamma)$, with $L$ clipped to $[1,8]$. We choose $\mu$ such that $\mathbb{E}[L]=6$ before clipping, meaning the model learns on average from $5.32$ loops during pretraining (Figure~\ref{fig:icl_loops_dist}).

Prior work optimizes the expected loss across recurrent depths \citep{zhu2025scaling,geiping2026scaling}. In contrast, we compute the cross-entropy loss only at the final iteration, training the model to achieve peak performance at the given $L$. Since our residual scaling in \ourModel{} depends on the number of iterations, this objective explicitly trains the model to use all iterations effectively.

\boldheading{Stage 1: Pretraining (500K steps).} We train for $500{,}000$ steps with batch size 64 on datasets containing 2-100 features, up to 10 classes, and up to 1{,}024 rows. Train/test split ratios are uniformly sampled from $[0.3,0.9]$. We use full backpropagation through all recurrent iterations. Optimization uses Muon \citep{jordan2024muon} with learning rate $8\times10^{-4}$ and momentum $0.9$. We use a $5{,}000$-step warmup followed by cosine decay over the final $10\%$ of training, gradient clipping at 1.0, weight decay 0.01.

\boldheading{Stage 2: Long-context continued pretraining (30K steps).} We continue training for $30{,}000$ steps on larger synthetic tasks with 1--2{,}000 features and up to $100{,}000$ rows, subject to a feature$\times$row budget of $10^6$. We use truncated backpropagation through time (TBPTT) \citep{geng2021training} with a window of $k=5$, propagating gradients only through the final five loops and detaching earlier activations. This reduces memory and computation while preserving parameter sharing across loops, allowing us to pretrain with a larger number of loops than would be feasible with full backpropagation under our GPU memory constraints. We use a learning rate of $10^{-4}$ and 400 warmup steps, with all other optimization settings kept the same as in Stage~1.

\boldheading{Stage 3: Early-exit gate (4K steps).} Starting from the Stage~2 checkpoint, we freeze the backbone and train the early-exit gate for $4{,}000$ steps with the same setting as Stage~2. The gate is trained on per-loop prediction losses: it continues if future iterations improve the loss and exits otherwise.

The gate uses only training embeddings and makes its decision based on the training samples, without using information from the test set. When the gate probability exceeds a threshold $\lambda$, the recurrent loop terminates, and the decoder is then run to obtain predictions. Notably, we train the early-exit gate with a maximum of 16 loops, which is twice the pretraining limit.

\section{Experiments}
\label{experiments}
In this section, we first conduct a series of ablation studies on \ourModel{} at computationally feasible intermediate pre-training checkpoints and investigate the model's performance across different layers and computational budgets (extensive results are in Appendix~\ref{app:experiment}). Finally, we evaluate \ourModel{} on \TabArena{} and \Talent{} benchmarks and compare its performance against state-of-the-art models.

\subsection{\rqloopstack{}}

\boldheading{Setup.} We evaluate checkpoints from Stage 1 ($100$K steps). We study classification performance averaged across $1{,}000$ \textit{synthetic tasks} sampled from the prior and across $10$ folds of the $54$ \textit{development set} used in \textit{TabICLv2} and \textit{TabPFNv2} \citep{hollmann-nature25a}
, without any ensembling. We use up to $2{,}048$ rows per dataset to match the pretraining regime, following~\citep{qu2026tabiclv2}.

We compare a fixed-depth TFM comprising six \textit{Stacked blocks} (\legendline{Fuchsia}{solid}), matching the approximately $\mathbb{E}[L]=5.21$ loops used during pretraining. As recurrent baselines, we consider \ourModel{} with $L=6$ with different residual scaling schedules $\alpha \in \{\textcolor{Aqua}{1}, \textcolor{Violet}{\sfrac{1}{L}}, \textcolor{Gray}{\sfrac{1}{\sqrt{L}}}\}$. Additionally, we evaluate a variant without distributional input encoding (\textit{w/o dist. conditioning}; \legendline{Cornflower}{dashed}).

\begin{figure}[tpb]
    \centering
    \shortstack[c]{
    \small \quad Training Stability\\[-0.1em]
    \includegraphics[height=0.15\textheight]{
        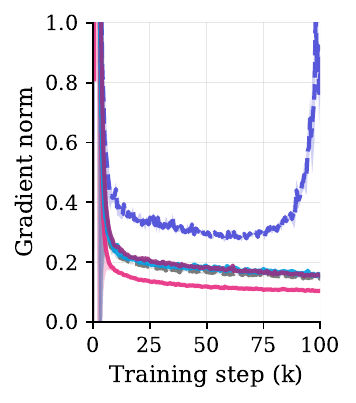
        }
    }
    \hfill
    \makebox[0.24\textwidth][c]{
        \shortstack[c]{
            \small \quad Development set\\[-0.1em]
            \includegraphics[height=0.15\textheight]{
                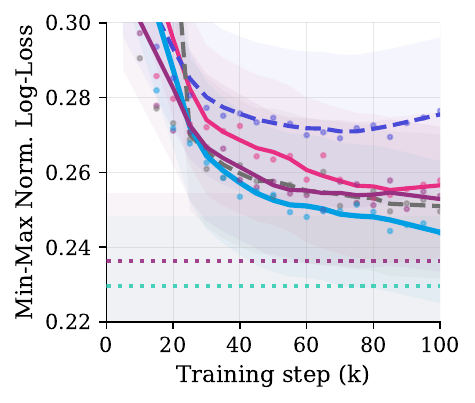
            }
        }
    }
    \hfill
    \makebox[0.24\textwidth][c]{
        \shortstack[c]{
            \small \quad   Synthetic priors\\[-0.1em]
            \includegraphics[height=0.15\textheight]{
                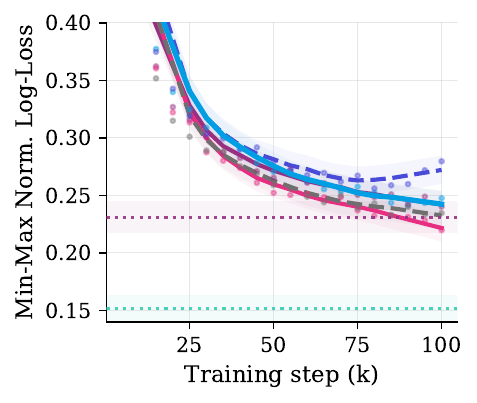
            }
        }
    }
    \hfill
    \includegraphics[height=0.15\textheight]{
        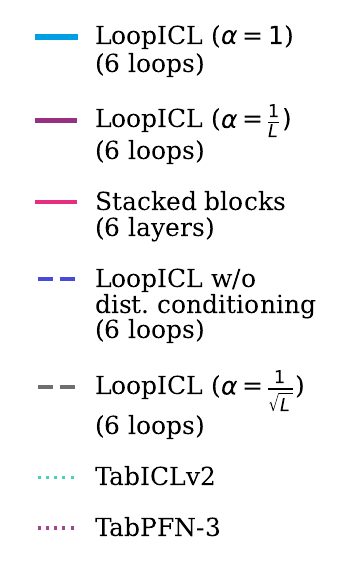
    }
    \caption{ \textbf{Ablation study.} Training stability (gradient norm) and predictive performance (normalized log loss averaged across tasks) of recurrent and non-recurrent model variants.}
    \label{fig:ablation}
\end{figure}

\boldheading{Pretraining stability.} The left panel of Figure~\ref{fig:ablation} compares the gradient norms as a proxy to training stability. Comparing the stacked blocks \legendline{Fuchsia}{solid} and the recurrent model \legendline{Aqua}{solid}, we observe that recurrent models are more challenging to train, exhibiting generally larger gradient norms. More interestingly, comparing the \legendline{Aqua}{solid} and \legendline{Cornflower}{dashed} curves shows that our \textit{distributional encoding} not only reduces gradient norm and, thus, improves training stability, but also achieves a better predictive performance across both \textit{synthetic priors} and the \textit{development set}.

\finding{Recurrence makes optimization challenging due to greater gradient instability.}
\finding{Distributional conditioning increases training stability and predictive performance.}

\boldheading{Performance.} The middle and right panel of Figure~\ref{fig:ablation} compares the stacked (\legendline{Fuchsia}{solid}) with recurrent variants (\legendline{Aqua}{solid}, \legendline{Violet}{dashed}, \legendline{Gray}{dashed}) on \textit{synthetic prior} and \textit{development set} tasks. On the \textit{synthetic priors} tasks, the stacked model achieves lower in-distribution loss, potentially reflecting its larger parameter count and greater capacity.\footnote{We note, that the fixed-depth may require additional training to undergo grokking to transition from memorization to generalization.
} In contrast, the recurrent variants achieve lower loss on the \textit{development set} tasks, indicating better generalization to unseen tasks. We also observe stronger generalization to large context sizes for our recurrent models (see Appendix~\ref{app:length_generalization})

\boldheading{Residual Scaling.} Finally, we compare different residual scaling strategies and revisit Figure~\ref{fig:ablation}. Overall, residual scaling has only a minor effect on the gradient norms; however, although we use a fixed $L$, performance on the \textit{development set} tasks varies. This raises the question of how residual scaling impacts stability during looping, which we study next.

\subsection{\rqrefine{}}

The most intriguing ability of looped models is that we can increase computational depth at inference time. Residual scaling can be used to controls this ability. Thus, we study how different scaling schemes impact generalization performance and internal representations during refinement.

\boldheading{Setup.} We continue pretraining for our looped models with different values of $\alpha$ for $10{,}000$ more steps in the Stage 2 setting to enable longer-context processing. We use all available samples in the \textit{development set} tasks and use the setup from the previous experiment otherwise.

\begin{figure}[htb]
    \centering
    \hfill
            \includegraphics[height=0.13\textheight]{
            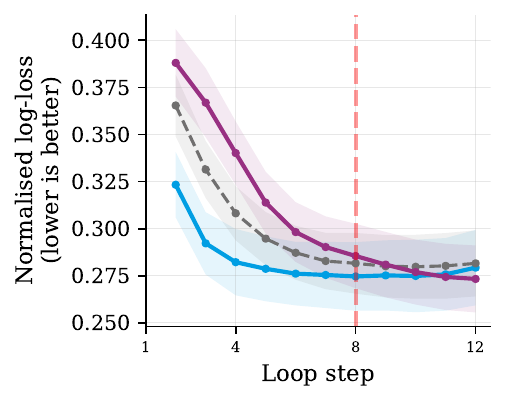
        }
                \includegraphics[height=0.13\textheight]{
            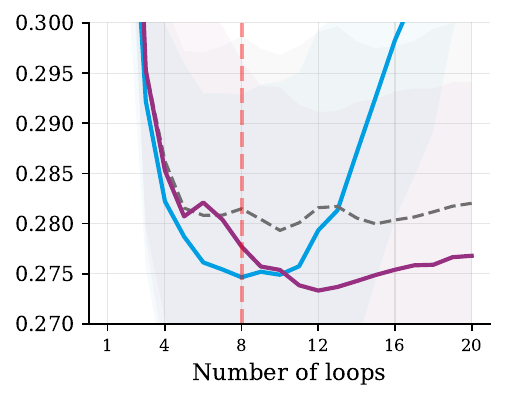
        }
                \includegraphics[height=0.13\textheight]{
            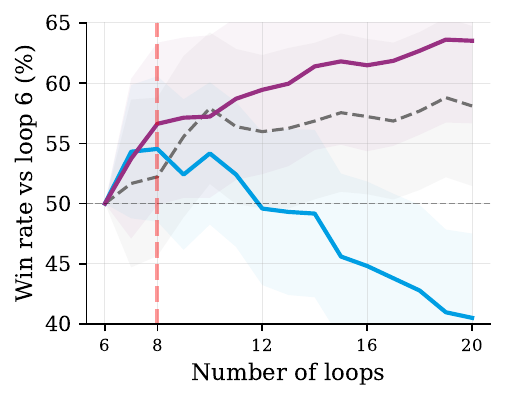
        }
        \includegraphics[height=0.10\textheight, trim=0 -10mm 0 00mm]{
            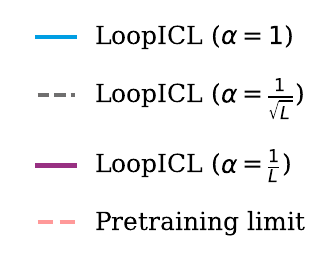
        }
     \caption{\textbf{Residual scaling improves extrapolation.} Left: per-iteration norm. log loss for $L=12$. Middle: norm. log loss for $L \in [1, 20]$. Right: win rate relative to performance at $L=6$, for $L\geq6$.}
    \label{fig:extrapolation}
\end{figure}

\boldheading{Impact on performance.} We first study how performance evolves during looping and evaluate predictions at each loop step up to $L=12$ (see Figure~\ref{fig:extrapolation}).

Aggressive fixed residual scaling ($\alpha=1$, \legendline{Aqua}{solid}) leads to slower improvements in performance, however, adaptive scaling ($\alpha=\sfrac{1}{\sqrt{L}}$, \legendline{Fuchsia}{dashed}) may achieve better performance at a higher loop count.

This suggests that, in settings such as early exit, \ourModelNoRS{} may be a preferable choice, as it can achieve stronger performance at earlier loop steps.

In the middle and right panel of Figure~\ref{fig:extrapolation}, we compare performance across increasing values for $L$.\footnote{Note that for $\alpha=1$ the value of $L$ does not change model behavior.}
\ourModelNoRS{} performs best within the range of loop counts sampled during pretraining, but degrades with high loop counts.

In contrast, adaptive residual scaling using \ourModelDefualtRS{} steadily improves performance with more loops even extrapolates beyond the pre-training limit.

However, as the scaling factor depends on the number of loops which, in practice, must be determined in advance. With a fixed scaling $\alpha=1$, $L$ does not need to be set in advance and the model can exit anytime.

\finding{Residual scaling enables stable extrapolation to loop counts beyond pretraining.}

\begin{figure}[htb]
    \centering
            \includegraphics[height=0.14\textheight]{
            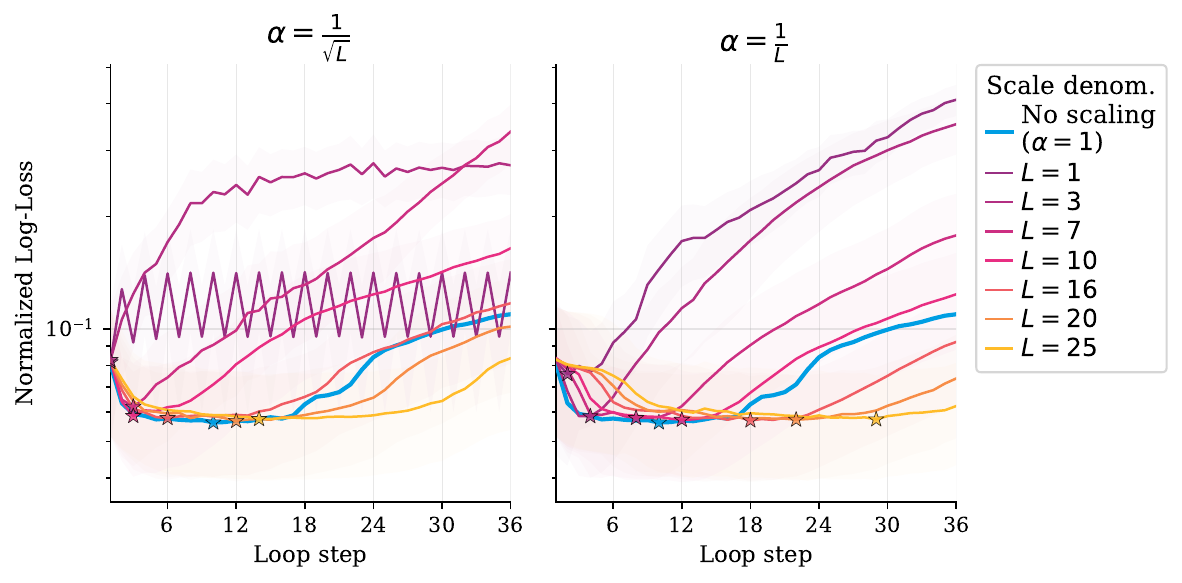
        }
    \hfill
                \includegraphics[height=0.14\textheight]{
            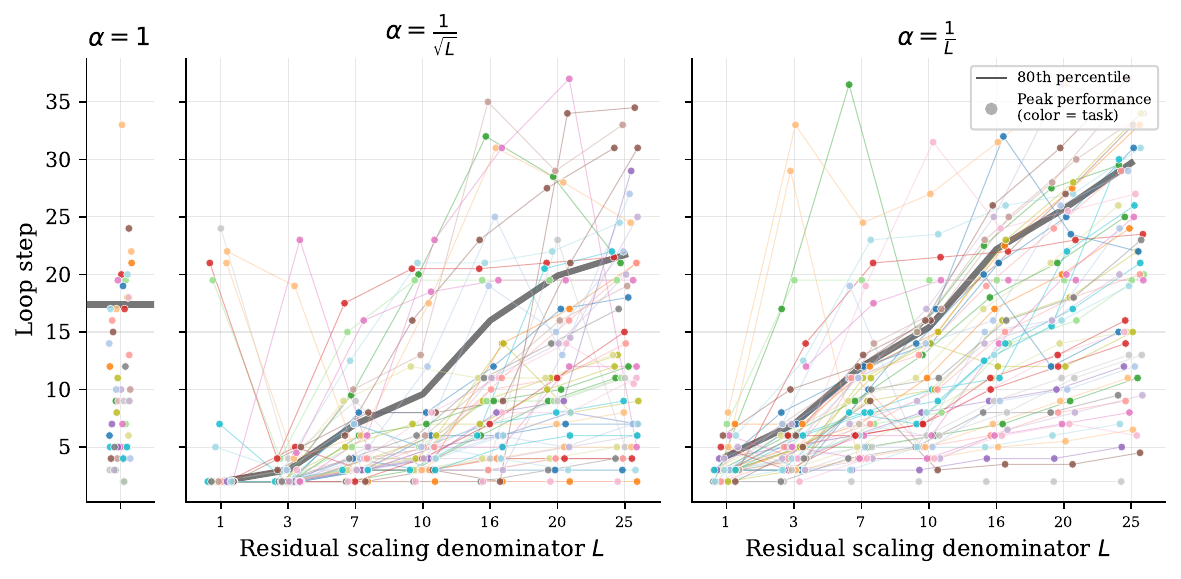
        }
    \caption{\textbf{Residual scaling ablation.} Left: Performance across loop steps for different $L$ and residual scaling schemes. Stars indicate the optimal loop step. Right: Optimal loop steps across datasets, showing dataset-dependent computation and the effect of $L$ on budget allocation.}
    \label{fig:ablation_on_residuals}
\end{figure}

\boldheading{Impact on model behavior.} Next, we characterize convergence behavior. For this we vary the value of $L$ (impacting residual scaling) and report per-loop performance, including the best-observed performance, across 1-36 loops in the left panel of Figure~\ref{fig:ablation_on_residuals}.

Overall, smaller values of $L$ lead to faster convergence, i.e. the best iteration, while increasing the number of iterations can cause performance divergence. Interestingly, for $\alpha = \sfrac{1}{\sqrt{L}}$ and $L=1$, we observe substantial fluctuations in performance. This behavior is reminiscent of the oscillatory or non-convergent dynamics that can arise in iterative fixed-point methods, which have recently been studied in the context of looped Transformers \citep{movahedi2026fixed}. As $L$ increases, convergence becomes slower, while the region of near-optimal performance (error $\leq 0.001$) gets broader. Also, we note that the best iteration not necessarily coincides with $L$, highlighting a possibility for further tuning.

The right panel of Figure~\ref{fig:ablation_on_residuals} shows that the optimal number of loop steps is task-dependent.
For the $\alpha=1$ variant, on average, approximately half of the datasets converge within $8$ loops, while the remaining datasets require more iterations. For $\alpha=\sfrac{1}{\sqrt{L}}$ and $\alpha=\sfrac{1}{L}$ the optimal loop count varies with $L$, indicating that the model adapts its computation to the given budget: the effective number of iterations required to reach near-optimal performance increases with $L$. In particular, $\alpha=\sfrac{1}{L}$ provides an effective allocation as the optimal number of loop steps approximates $L$.

\finding{With $\alpha=1$, the model optimally scales updates across all loop steps within the pretraining range. Adaptive scaling using $\alpha=\sfrac{1}{L}$ enables budget-aware computation, achieving near-optimal performance around $L$ loop steps, including beyond the pretraining range.}

\begin{figure}[tbp]
    \centering
     \includegraphics[height=0.14\textheight]{
            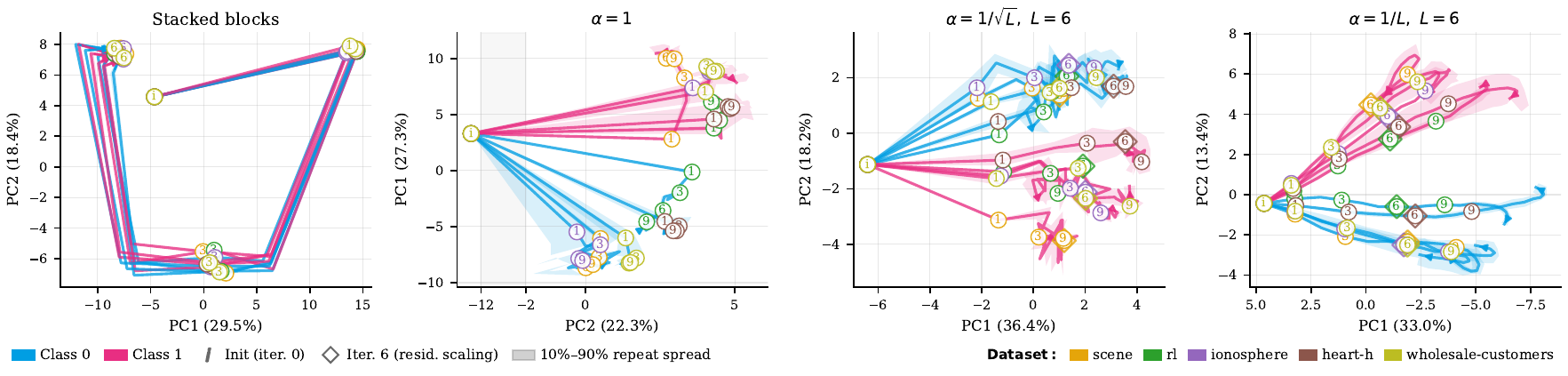
        }
     \caption{\textbf{Latent representation trajectories.} PCA of embeddings across loops for binary tasks. Residual scaling with $\alpha=\sfrac{1}{L}$ induces structured and uniform progress toward separating classes.}
    \label{fig:latent_pca}
\end{figure}

\boldheading{Internal representations.} Finally, we analyze how the internal representations and performance evolve at each loop step (see Appendix~\ref{app:representation_analysis} for further details). Figure~\ref{fig:latent_pca} shows the evolution of the embedding space across loop steps. We track internal embeddings for five binary classification tasks from the \textit{development set} across $L=6$ loops and visualize the first two PCA components.

Starting from the right-most panels, $\alpha=\sfrac{1}{\sqrt{L}}$ and $\alpha=\sfrac{1}{L}$ exhibit a very structured and similar behavior. PC2 primarily captures class separation across the datasets, while PC1 is strongly correlated with the loop step. The model continuously changes its internal state with more loops. Interestingly, for $\alpha=\sfrac{1}{L}$, the loop steps are encoded almost uniformly resulting in linear trajectories, suggesting that $\alpha=\frac{1}{L}$ induces a controlled iterative refinement.

For $\alpha=1$ PC1 ($27.3\%$ explained variance) also captures class separation, however, the evolvement of representation through PC2 is less systematic, suggesting that the representations quickly organize according to the target classes. The left panel shows the fixed-depth model as a reference (more details in Appendix~\ref{app:pca_analysis}).

\finding{$\alpha=\sfrac{1}{L}$ yields more stable and uniform encoding of loop steps, leading to smoother latent representation trajectories and more consistent iterative refinement.}

\subsection{\rqperf{}}

Finally, we study performance on \TabArena{}~\citep{erickson2025tabarena} and \Talent{}~\citep{liu2025talent}.

We compare against state-of-the-art TFMs and report improvability vs. inference costs in Figure~\ref{fig:elo_improvability} (detailed results are in Appendix~\ref{app:performance}). We evaluate \ourModel{} using an ensemble of 8 members. We choose $\alpha=1$ to apply our early exit gate to showcase both ways to control the performance-cost tradeoff: by fixing the number of loops $L$ we define the compute budget to be used and with the early-exit threshold $\lambda > 0$ we specify a confidence threshold required to justify another loop.

Specifying $L$, the performance of \ourModel{} improves with more loops, peaking at around $L=12$ loop steps, competitive with \TabICLvTwo{}. \TabICLvTwo{} uses 12 ICL blocks and hidden dimensions of the same size as \ourModel{}, thus, the dominant FLOPs cost is the same (runtime measurements might differ due to different hardware (see caption of Figure \ref{fig:elo_improvability}); \ourModel{} incurs only a minor overhead from additional cheap row-interaction and set-transformer iterations. Notably, \ourModel{} achieves this at $88.2\%$ fewer parameters.

More importantly, these results demonstrate that our model robustly extrapolates beyond its pretraining limit on real-world tasks and operates effectively at roughly twice the average number of loop steps observed during pretraining.

\begin{figure}[bt]
\centering
\begin{subfigure}[b]{0.48\linewidth}
    \centering
    \small{{TabArena}}
    \includegraphics[width=\linewidth]{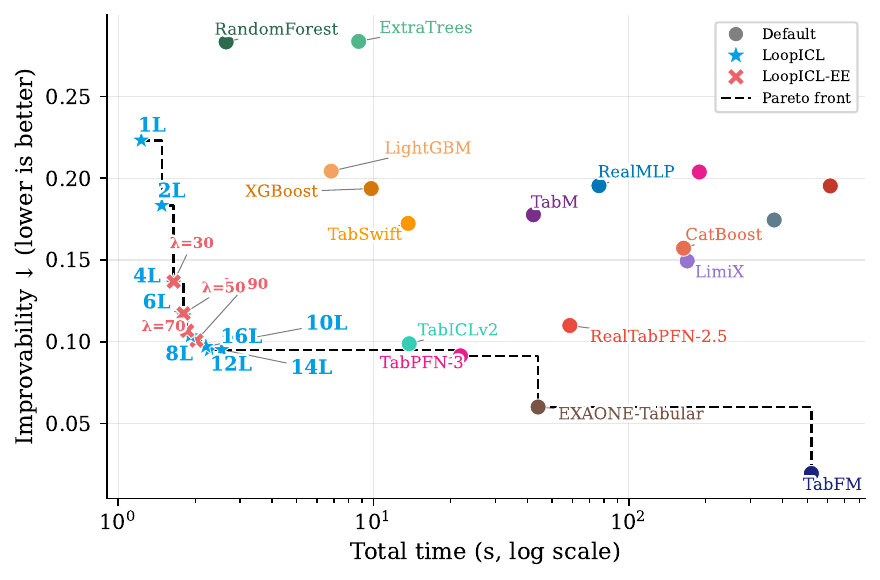}
\end{subfigure}
\hfill
\begin{subfigure}[b]{0.48\linewidth}
    \centering
    \small{{TALENT}}
    \includegraphics[width=\linewidth]{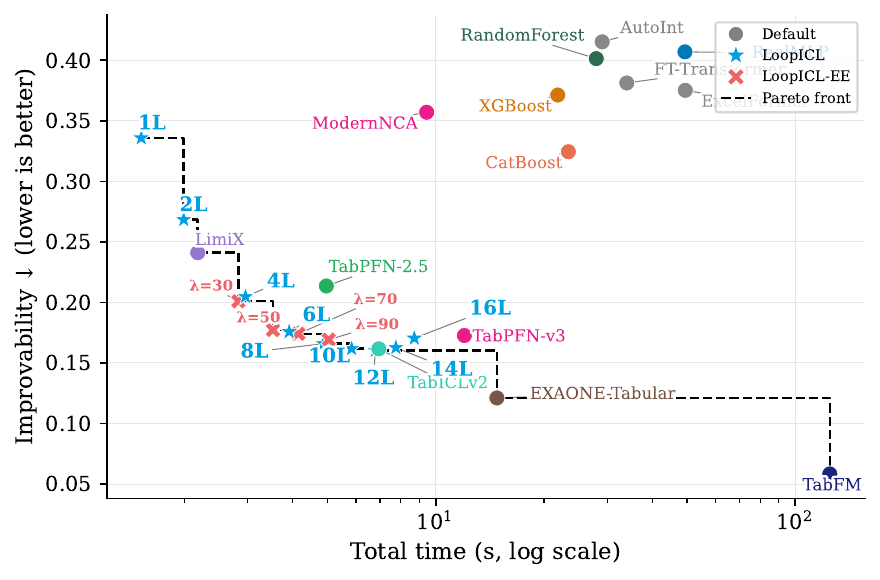}
\end{subfigure}
\vspace{-0.5em}
\caption{Improvability (lower is better) measures the relative
error gap to the best method, averaged across datasets. Time is training + inference. For \Talent{}, all models are evaluated on an A100 GPU under identical conditions, enabling a direct runtime comparison. For \TabArena{}, competitor runtimes are taken from the published benchmark and may reflect different hardware; these results should not be used for direct runtime comparisons of \ourModel{} with other baselines.}
\label{fig:elo_improvability}
\end{figure}

Results using the early-exit gate (\ourModel{}-EE) confirm this as a reliable alternative to relying on a pre-defined, fixed number of loop steps.

\begin{figure}[bt]
\centering
\begin{subfigure}[b]{0.42\linewidth}
    \centering
    \includegraphics[width=\linewidth]{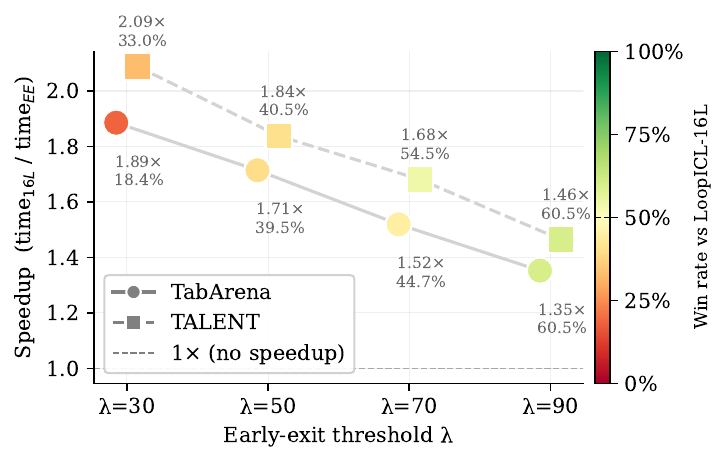}
\end{subfigure}
\hfill
\begin{subfigure}[b]{0.28\linewidth}
    \centering
    \footnotesize{TabArena}\\[2pt]
    \includegraphics[width=\linewidth]{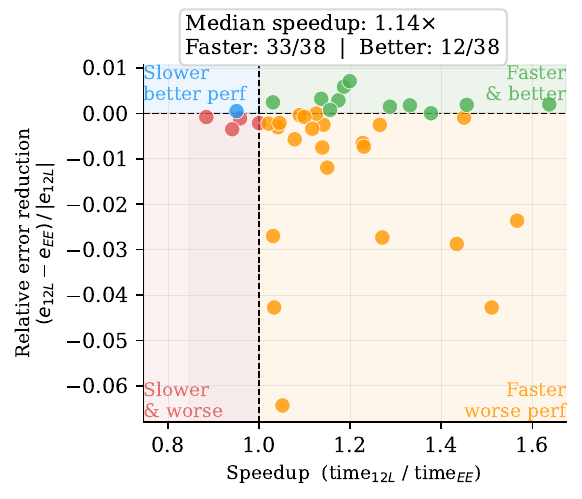}
\end{subfigure}
\hfill
\begin{subfigure}[b]{0.28\linewidth}
    \centering
    \footnotesize{TALENT}\\[2pt]
    \includegraphics[width=\linewidth]{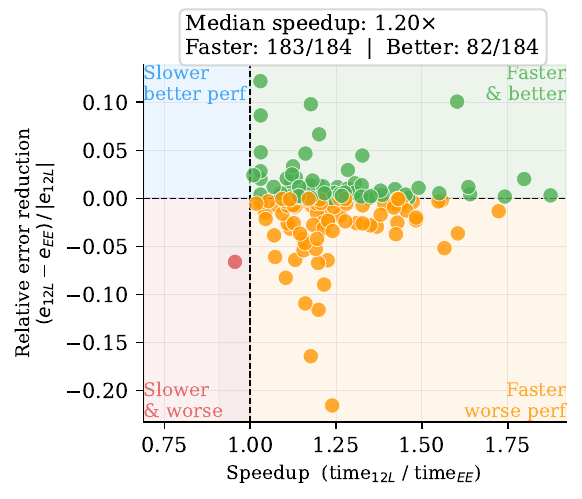}
\end{subfigure}
\caption{\textit{Left:} Median speedup of \ourModel{}-EE over the full 16-loop baseline across different $\lambda$; color indicates the fraction of datasets where EE wins. \textit{Center / Right:} Per-dataset speedup vs.\ relative error reduction for $\lambda{=}90$ on \TabArena{} and \Talent{} , compared to the 12-loop baseline (the best-performing fixed-loop configuration). Each point is one dataset; $x>1$ means EE is faster, $y>0$ means EE is more accurate. Points in the green quadrant achieve both simultaneously.}
\label{fig:ee_speedup_vs_perf}
\end{figure}

The left panel in Figure~\ref{fig:ee_speedup_vs_perf} reports the median speedup relative to the 16-loop baseline (i.e., early exit disabled, 8 ensemble members each running 16 loops), which serves as the upper bound on the number of loops executed. As $\lambda$ increases, the model exits more conservatively, yielding better performance at the cost of lower speedup. At $\lambda{=}90$ we observe both a speedup and a win-rate gain over the 16-loop baseline on both benchmarks.

The center and right panels in Figure~\ref{fig:ee_speedup_vs_perf} compare $\lambda{=}90$ against the best-performing fixed-loop configuration (12 loops). On \TabArena{}, EE is faster on 33 out of 38 datasets, and 12 out of 38 datasets gain both speed and accuracy simultaneously. Some datasets do incur a performance loss, though the maximum relative error increase remains within $6\%$. On \Talent{}, we observe higher overall speedup and win rate; however, the magnitude of both gains and losses is larger, with maximum relative error gains and losses reaching approximately $10\%$ and $20\%$, respectively.

\section{Conclusion}
\label{conclusion}

Our results revisit the central question: does tabular ICL require \emph{more parameters}, or simply \emph{more computation}? \ourModel{} suggests the latter. A single shared block, applied recurrently, matches substantially larger fixed-depth TFMs on \TabArena{} and \Talent{}. This also provides evidence that tabular foundation models may rely more on learning iterative task-solving algorithms than on memorizing and recalling knowledge, as is common in language models.
Across RQ1--RQ3, we find that weight sharing improves generalization, recurrent computation enables stable iterative refinement even beyond the pretraining depth, and the number of loops can adapt to dataset difficulty at inference time. \textbf{These results suggest that some benefits of model scaling may arise from increased computational depth rather than parameter count, motivating a simple alternative: instead of growing the parameter budget, grow the loop budget.}

\boldheading{Limitations and future directions.} The current model supports classification tasks; extending it to regression tasks is left for future work.
Future work should scale \ourModel{} in both model size and the number of recurrent iterations, for example by repeating $N$ layers rather than a single layer. Studying scaling laws for tabular foundation models is another promising direction. Finally, inspired by low-rank structure observed in KV caching across iterations in looped language models \citep{vendrell2026memory, neill2026looped}, it would be interesting to investigate whether similar structure exists in the tabular domain and can be exploited for more efficient inference.

%\subsection*{AI use statement}
%In this work, we used generative AI tools for polishing text and identifying errors. AI assistance was used during the development of the project for documentation, code cleaning and editing, development of small helper functions, and plotting. We did not use generative AI tools to propose or refine the method or hypotheses, or to generate data sets. Tasks related to mathematical proofs, translation and qualitative data analysis are not applicable to this work.  We have reviewed all AI-assisted work and take full responsibility for the final content of this work, including text, claims, and artifacts produced with the aid of generative AI.

\subsection*{Implementation Code}
The implementation code is available at \url{https://github.com/amirbalef/loopicl}.

\subsubsection*{Acknowledgments}
This research has been funded by the Federal Ministry of Research, Technology and Space of Germany and the state of North Rhine-Westphalia as part of the Lamarr Institute for Machine Learning and Artificial Intelligence. Additionally, A. Balef thanks the International Max Planck Research School for Intelligent Systems (IMPRS-IS).

\bibliography{strings, references,lib,proc,myproc}
\bibliographystyle{conference}

\appendix
\section{Architecture Details}
\label{app:architecture}

\subsection{Input Encoding}
Raw features are standardized per column using training-row statistics, with missing values imputed by the column mean.
Following \citet{qu2026tabiclv2}, the model applies \emph{cyclic feature grouping}: each feature is concatenated with its $G{-}1$ cyclically-shifted neighbors ($G{=}3$), forming local feature groups projected to $E$-dimensional cell embeddings, yielding the initial cell stream $\mathbb{R}^{R \times C \times E}$.

\textbf{Label encoding.}
Training-row class labels are mapped through a trainable orthogonally-initialized embedding \citep{grinsztajn2026tabpfn} and added to training-row cells.

\textbf{Distribution conditioning.}
Three signals characterizing the training distribution are computed once and broadcast across all rows.
\begin{itemize}
\item \textbf{Marginal histogram.} Adaptive-quantile soft log-histograms (32 bins) with four distributional moments are projected column-wise via a small MLP. 
\item \textbf{Discriminative histogram.} For \emph{classification}, per-class log-histogram residuals (class minus marginal) are projected column-wise. 
\item \textbf{Fourier quantile encoder.} The empirical rank of each cell within its column is Fourier-encoded (8 frequencies) with a binary OOD flag and projected per-cell. 
\end{itemize}
All conditioning MLPs have zero-initialized output layers, making them initially inactive \citep{bachlechner2021rezero}. This prevents untrained conditioning heads from disrupting the backbone during pretraining and protects pretrained representations during regression continued pretraining \citep{zhang2023adding}. The conditioning pathways then gradually activate as their weights move away from zero, allowing the model to learn how strongly to rely on each signal.

\subsection{Recurrent Block}
Each recurrent iteration applies a single \emph{shared} axial block with weights tied across all $L$ loops. 
\emph{Axial} here means the block alternates attention over rows (within each column) and attention over columns (within each row), avoiding the quadratic cost of full joint attention over all cells.
For the \emph{cell stream} ($\mathbb{R}^{R \times C \times E}$):
\begin{enumerate}
    \item \textbf{Label re-injection.} Label embeddings are added to training-row cells via a zero-initialized projection.
    \item \textbf{Within-column attention.} An Induced Set Attention Block (ISAB) \citep{qu2025tabicl} processes each column independently over rows using $K{=}128$ learned inducing points ($H{=}8$, head dim 16).
    \item \textbf{Cross-column attention.} Self-attention with RoPE \citep{su2024roformer} processes each row over columns ($H{=}8$, head dim 16).
\end{enumerate}
A \textbf{readout} via cross-attention then projects cells into the \emph{row stream} ($\mathbb{R}^{R \times D}$) using $C_\ell{=}4$ learned CLS tokens per row.
The row stream is updated by a zero-initialized label injection followed by causally-masked ICL attention ($H{=}8$, head dim 64), allowing test rows to attend to training rows.
All sub-blocks use sandwich normalization (pre- and post-RMSNorm) \citep{ding2021cogview}.

Each attention block includes a query-conditioned softmax scaling layer \citep{qu2026tabiclv2} that rescales queries before the dot product:
\begin{equation*}
  q \;\leftarrow\; q \cdot \gamma \cdot \delta, \qquad
  \gamma = 1 + f\!\left(\log\tfrac{n}{n_\text{ref}}\right), \qquad
  \delta = 1 + \tanh\!\left(g(q)\right),
\end{equation*}
where $\gamma$ is a per-head temperature correction, $\delta$ is a per-query modulation, and $f$, $g$ are small zero-initialized MLPs.
We introduce $n_\text{ref}{=}512$ to prevent unbounded output growth that can cause instability at longer sequence lengths \citep{pfeiffer2026tabh2o}.

\subsection{Output Head} Inspired by \citet{grinsztajn2026tabpfn}, an attention-based retrieval decoder \citep{arbel2026equitabpfn} accumulates per-class attention mass between test-row queries and training-row keys via one-hot label weighting \citep{koshil2025context}, averaged across heads to produce class log-probabilities.

\subsection{Early-Exit Gate}
\label{app:earlyexit_gate}
After each recurrent iteration $t$, a lightweight gate decides whether to stop or continue.  If it exits at step $t$, the decoder runs once at that step; earlier iterations only update the shared representations. The gate has fewer than 20K ($19\,521$) parameters and runs on pooled row embeddings, so its overhead is negligible.

\paragraph{Gate inputs.}
The gate takes a 43-dimensional feature vector $\mathbf{f}_t$ formed by concatenating five signals:

\begin{center}
\small
\begin{tabular}{llc}
\toprule
Signal & Description & Dim \\
\midrule
$\mathbf{c}_t$ & Compressed mean state & 8 \\
$\Delta_t = \mathbf{c}_t - \mathbf{c}_{t-1}$ & Change since last loop & 8 \\
$\mathrm{pe}(t)$ & Sinusoidal loop-depth encoding & 16 \\
$[\lambda_{t-1}, \lambda_{t-2}, \lambda_{t-3}]$ & Last three gate outputs & 3 \\
$\mathrm{pe}(\log N)$ & Sinusoidal dataset-size encoding & 8 \\
\midrule
\multicolumn{2}{l}{Total} & 43 \\
\bottomrule
\end{tabular}
\end{center}

The compressed state $\mathbf{c}_t \in \mathbb{R}^8$ is computed by passing each row embedding through a small MLP (512\,→\,32\,→\,8, ReLU, no bias) and averaging over rows.  The change signal $\Delta_t$ is near zero when representations have stabilised, providing a natural convergence cue.

\paragraph{Decision network.}
A two-layer MLP (43\,→\,64\,→\,1) with sigmoid output maps $\mathbf{f}_t$ to an exit probability:
\[
  \lambda_t = \sigma\!\left(\mathrm{MLP}_\phi(\mathbf{f}_t)\right) \in (0,1).
\]
The output layer is zero-initialised, so $\lambda_1 \approx 0.5$ at the start of training.

\paragraph{Inference-time exit rule.}
We accumulate a running exit probability
\[
  F_t = 1 - \prod_{s=1}^{t}(1 - \lambda_s)
\]
and stop at the first loop $t$ where $F_t > q$ (default $q = 0.5$), following \citet{zhu2025scaling}.  A per-sample variant exits each example independently once its $F_t > q$.

\paragraph{Training.}
The backbone is frozen; only the gate is trained.  The oracle label
\[
  w_t = \sigma\!\bigl(50\,(I_t - 0.005)\bigr), \qquad I_t = \mathcal{L}_t - \min_{s > t}\mathcal{L}_s,
\]
encodes whether continuing past loop $t$ reduces the loss.  We train with weighted binary cross-entropy, upweighting steps where early exit causes large regret ($r_t = (\mathcal{L}_t - \min_s \mathcal{L}_s)^+$):
\[
  \mathcal{L}_{\mathrm{gate}}
    = \frac{1}{T}\!\left[\sum_{t=1}^{T-1}(1 + r_t)\,\mathrm{BCE}(\lambda_t,\, 1 - w_t)
      \;+\; \mathrm{BCE}(\lambda_T,\, 1)\right].
\]
To discourage late exits we add a gap penalty $\mathcal{L}_{\mathrm{gap}} = 0.02\,\mathbb{E}[\tilde{g}(\bar{t} - t^*)^2]$, where $\tilde{g}(\delta) = 2\delta$ for $\delta > 0$ and $\delta$ otherwise (late exits penalised twice as heavily).  The total loss is $\mathcal{L} = \mathcal{L}_{\mathrm{gate}} + \mathcal{L}_{\mathrm{gap}}$.

\subsection{Parameter Counts}
Table~\ref{tab:arch_params} lists per-component parameter counts. In total, the model has ${\sim}3.25$M parameters with ${\sim}2.7$M parameters in the recurrent block.

\begin{table}[h]
\centering
\caption{Parameter counts for \ourModel{} ($E{=}128$, $D{=}512$).}
\label{tab:arch_params}
\small
\begin{tabular}{lr}
\toprule
\textbf{Component} & \textbf{Parameters} \\
\midrule
\multicolumn{2}{l}{\textit{Input encoding}} \\
Cell embedding                                  &         512 \\
Label encoder                                   &       1{,}280 \\
Log-histogram conditioner                       &      14{,}848 \\
Discriminative histogram conditioner            &      14{,}592 \\
Fourier quantile encoder                        &       9{,}472 \\
Init readout                                  &     131{,}712 \\
\midrule
\multicolumn{2}{l}{\textit{Shared axial block (tied across $L$ loops)}} \\
Within-column attention (ISAB)                  &     290{,}384 \\
Cross-column attention                          &     142{,}160 \\
ICL transformer block                           & 2{,}140{,}928 \\
Auxiliary (label inject, CLS, norms)            &      83{,}456 \\
\midrule
\multicolumn{2}{l}{\textit{Output}} \\
Output norm                                     &         512 \\
Retrieval decoder                               &     427{,}392 \\
\midrule
\textbf{Total}                                  & \textbf{3{,}257{,}248} \\
\bottomrule
\end{tabular}
\end{table}

\section{Inference Details}
\label{app:inference}

\paragraph{Preprocessing.}
\ourModel{} applies a fixed preprocessing pipeline fitted on training rows and reapplied consistently to test rows:
\begin{enumerate}
    \item \textbf{Feature encoding.} Categorical columns are ordinal-encoded (unknown test categories $\to -1$); numeric missing values are mean-imputed from training statistics.
    \item \textbf{Constant feature removal.} Features with only one unique training value are dropped.
    \item \textbf{Z-score standardization.} All features are standardized and clipped to $[-100, 100]$.
    \item \textbf{Outlier clipping.} A two-stage z-score clipper (threshold 4.0) first identifies and removes outliers, re-estimates statistics on the cleaned data, then clips remaining values to $\pm 4\hat{\sigma}$ of the cleaned distribution.
\end{enumerate}

\paragraph{Ensemble inference.}
At test time, \ourModel{} runs $N{=}8$ ensemble members and aggregates their predictions.
Each member receives a differently transformed view of the data, varied along three axes:
\begin{itemize}
\item \textbf{Feature order.} Columns are permuted using Latin square patterns \citep{qu2026tabiclv2}, ensuring each feature appears in a distinct input position across members with no two members sharing the same ordering.
\item \textbf{Class permutation (classification).} Training labels are shuffled using a balanced permutation strategy \citep{grinsztajn2026tabpfn}: all unique class orderings are enumerated, deduplicated, and distributed evenly across members, removing positional bias towards specific class indices.
\item \textbf{Preprocessing.} Members cycle through four distribution normalizations: \emph{none}, Yeo-Johnson, quantile-to-$\mathcal{N}(0,1)$, and robust IQR scaling, to increase diversity.
\end{itemize}
A fingerprint feature---a per-row hash of the training context---is appended to each member's feature matrix \citep{grinsztajn2026tabpfn}, providing a unique row identifier per member.
For \emph{classification}, each member's logits are converted to probabilities via softmax and the $N$ probability vectors are averaged.

\paragraph{Many-class support.}
The model natively supports up to $T{=}10$ classes. For datasets with more than 10 classes, we apply a mixed-radix decomposition \citep{qu2026tabiclv2}: each class $c$ is encoded as a tuple of digits $(d_1, \ldots, d_L)$ in a mixed base, where each digit position $d_\ell \in \{0, \ldots, b_\ell{-}1\}$ defines a $b_\ell$-class sub-problem the model can solve natively.
Each digit position is run independently through the full ensemble, producing per-digit class probabilities.
Final class probabilities are obtained by multiplying the per-digit probabilities: $P(c) \propto \prod_\ell P(d_\ell{=}\mathrm{digit}_\ell(c))$.

\section{Pretraining Details}
\label{app:pretraining}

\paragraph{Recurrent loop schedule.} At each training step, the number of recurrent iterations $L$ is sampled from a log-normal Poisson distribution \citep{geiping2026scaling}: $\gamma \sim \mathrm{LogNormal}(\mu, \sigma^2)$ with $\sigma = 0.5$, then $L \sim \mathrm{Poisson}(\gamma)$, clipped to $[1, 8]$. The mean $\mu$ is chosen so that $\mathbb{E}[L] = 6$. This distribution concentrates mass near the default depth while maintaining a long tail, enabling generalization to more iterations at inference time \citep{schwarzschild2021can}. During training $L$ is sampled from a log-normal Poisson distribution ($\mu{=}5$, $\sigma{=}0.5$, max~8); at inference $L{=}6$ with loop-residual scaling $\alpha{=}\gamma/(L\sqrt{S})$.

\paragraph{Stage 1: Pretraining.} We train for 500{,}000 steps with a batch size of 64 on synthetic datasets generated by the \TabICLvTwo{} graph-based SCM prior \citep{qu2026tabiclv2}. Each dataset contains 2--100 features, up to 10 classes, and up to 1{,}024 rows, with train/test split ratios uniformly sampled in $[0.3, 0.9]$. Full backpropagation through all recurrent iterations is used. We use the Muon optimizer \citep{jordan2024muon} (lr $= 8 \times 10^{-4}$, momentum $= 0.9$) with an AdamW fallback (lr $= 3 \times 10^{-4}$, $\beta = (0.9, 0.95)$) for scalar and embedding parameters. A warmup-stable-decay (WSD) schedule applies a 5{,}000-step warmup followed by cosine decay over the final 10\% of training. We use gradient clipping (norm $= 1.0$), weight decay $= 0.01$, an exponential moving average (EMA) with decay $= 0.999$, and \texttt{bfloat16} mixed precision.

\paragraph{Stage 2: Long-context fine-tuning.} Starting from the Stage~1 checkpoint, we fine-tune for 30{,}000 steps on a larger-scale prior with 1--2{,}000 features and up to 100{,}000 rows (subject to a feature$\times$row budget of $10^6$), enabling the model to handle datasets far beyond its Stage~1 training distribution. To make this computationally feasible, we use TBPTT \citep{geng2021training} with a window of $k = 5$: gradients are propagated only through the last 5 recurrent iterations, and earlier activations are detached. This reduces memory proportionally to $k/L$, allowing inference-time depth to far exceed training-time depth \citep{geiping2026scaling}. We use a reduced learning rate of $10^{-4}$ with 400 warmup steps; all other optimizer settings are inherited from Stage~1.

\paragraph{Training compute.} We train \ourModel{} on B300 GPUs. Stage 1 takes 18 GPU-days, Stage 2 takes 10 GPU-days, and training the early-exit gate takes 1 GPU-day.

\section{Experiment Details}
\label{app:experiment}

\subsection{Length Generalization}
\label{app:length_generalization}

\begin{figure}[t]
\centering
\includegraphics[width=0.75\linewidth]{
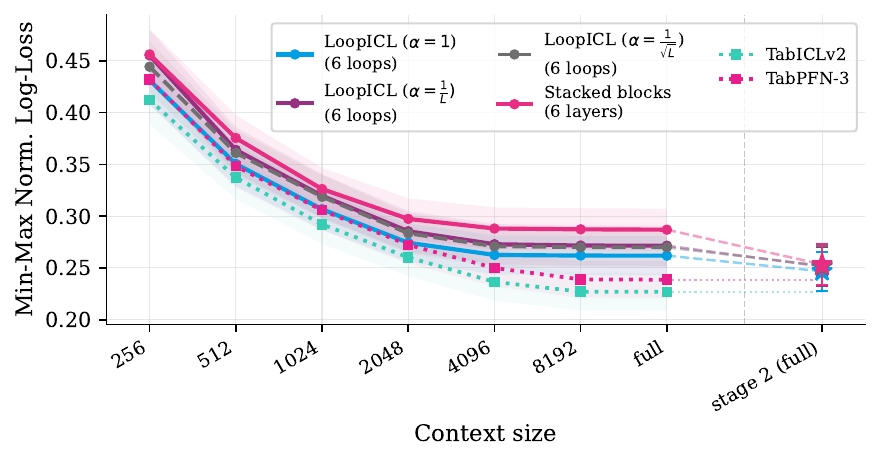
}
\caption{\textbf{Length generalization.} Recurrent models demonstrate stronger length generalization compared to the stacked model.}
\label{fig:length_gen}
\end{figure}

We also compare the ability of our models to scale with an increasing number of in-context examples. As shown in Figure~\ref{fig:length_gen}, the recurrent models scale slightly better than the stacked model, with the performance gap between the stacked (\legendline{Fuchsia}{solid}) and recurrent models (\legendline{Aqua}{solid}, \legendline{Violet}{dashed}, \legendline{Gray}{dashed}) widening as the context size increases. This effect is particularly pronounced for the recurrent model with $\alpha=\sfrac{1}{L}$, consistent with the findings of \citep{fan2025looped}, who show that looped Transformers with an adaptive number of steps can substantially improve length generalization.

We further evaluate the models after long-context training in Stage 2. As shown in Figure~\ref{fig:length_gen}, all models benefit from long-context training. Notably, after this additional training, the stack model nearly matches the performance of the recurrent models, substantially narrowing the gap observed with shorter-context training.

\subsection{Representation analysis}
\label{app:representation_analysis}
We analyze how embeddings evolve across loop iterations using three complementary measures: embeddings similarity, cross-step probing classifiers (logistic regression trained on step $i$, tested on step $j$), and class separation gap, following the approach of \citet{balef2026is}.

\begin{figure}[htpb]
\centering
\includegraphics[height=0.25\linewidth]{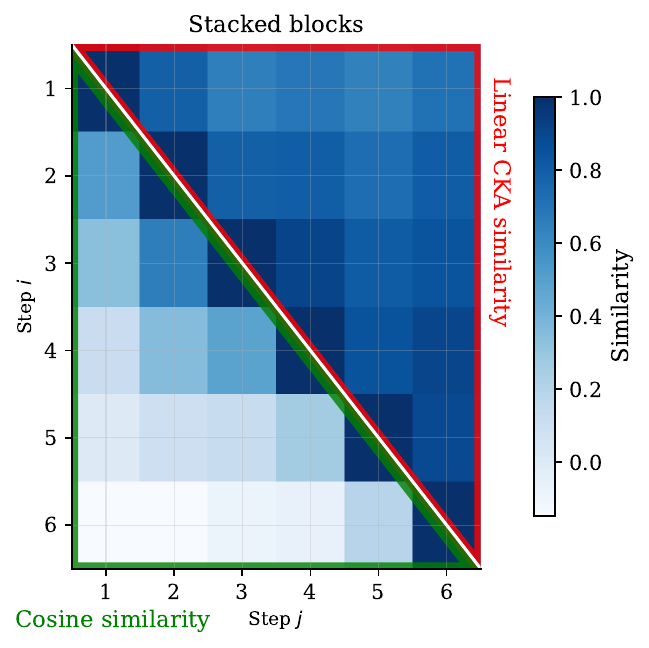}
\includegraphics[height=0.25\linewidth]{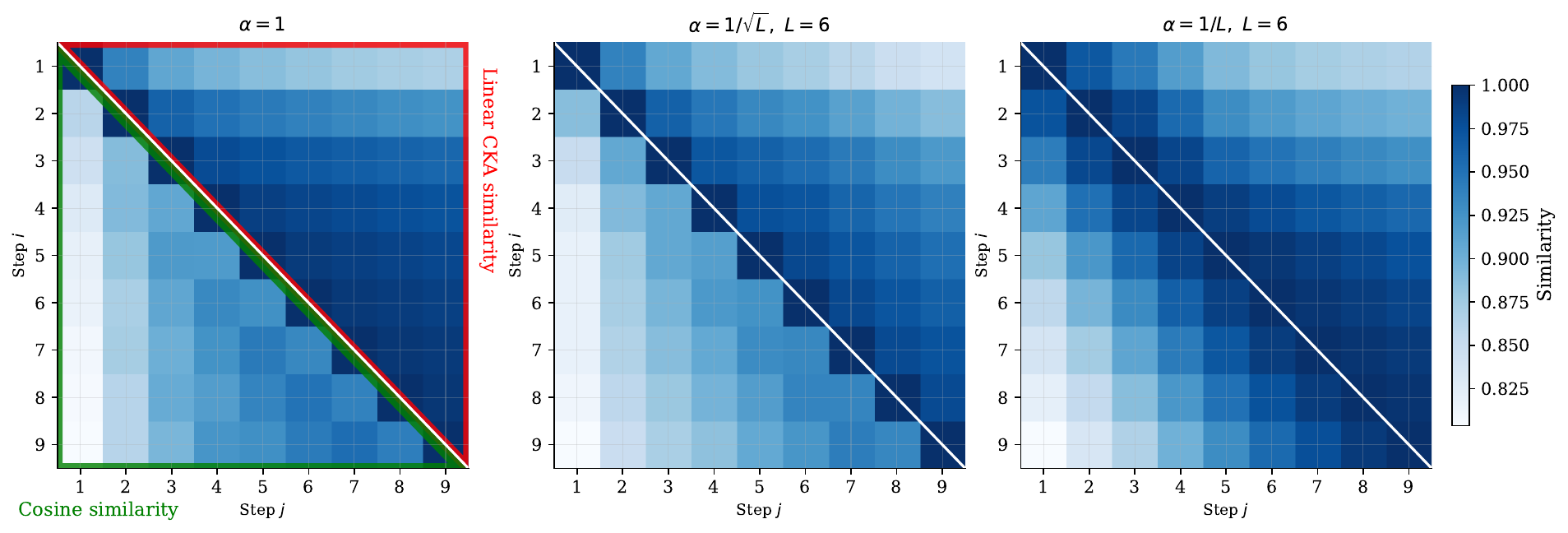}
\caption{Cross-step embedding similarity for stacked blocks (\textit{left}) and looped models (\textit{right}).
Each heatmap entry $(i,j)$ shows the cosine similarity (lower triangle) or linear CKA (upper triangle)
between embeddings at loop steps $i$ and $j$, averaged over all binary-classification datasets.}
\label{app:fig:embedding_similarity}
\end{figure}

\boldheading{Embeddings similarity.} As shown in Figure~\ref{app:fig:embedding_similarity}, stacked blocks exhibit low off-diagonal similarity, indicating that each block produces a qualitatively distinct representation. In contrast, looped models maintain high similarity across step pairs, reflecting the representational stability induced by weight tying.

\begin{figure}[htpb]
\centering
\includegraphics[height=0.25\linewidth]{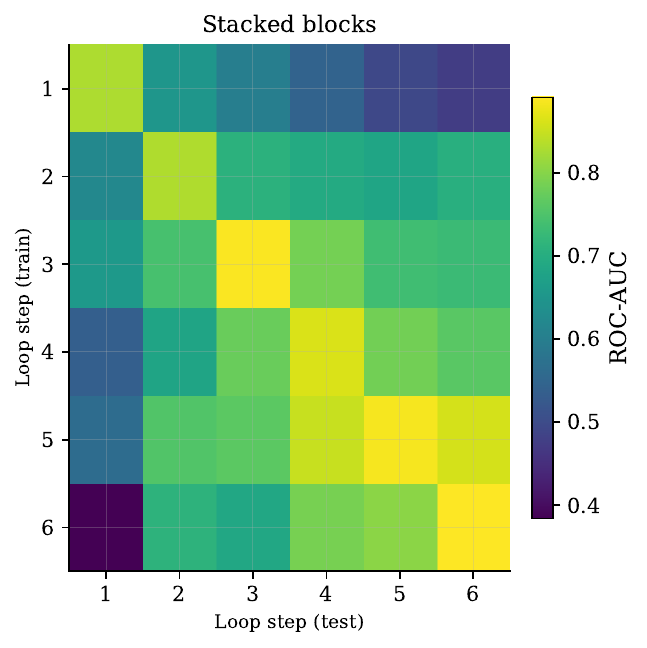}
\includegraphics[height=0.25\linewidth]{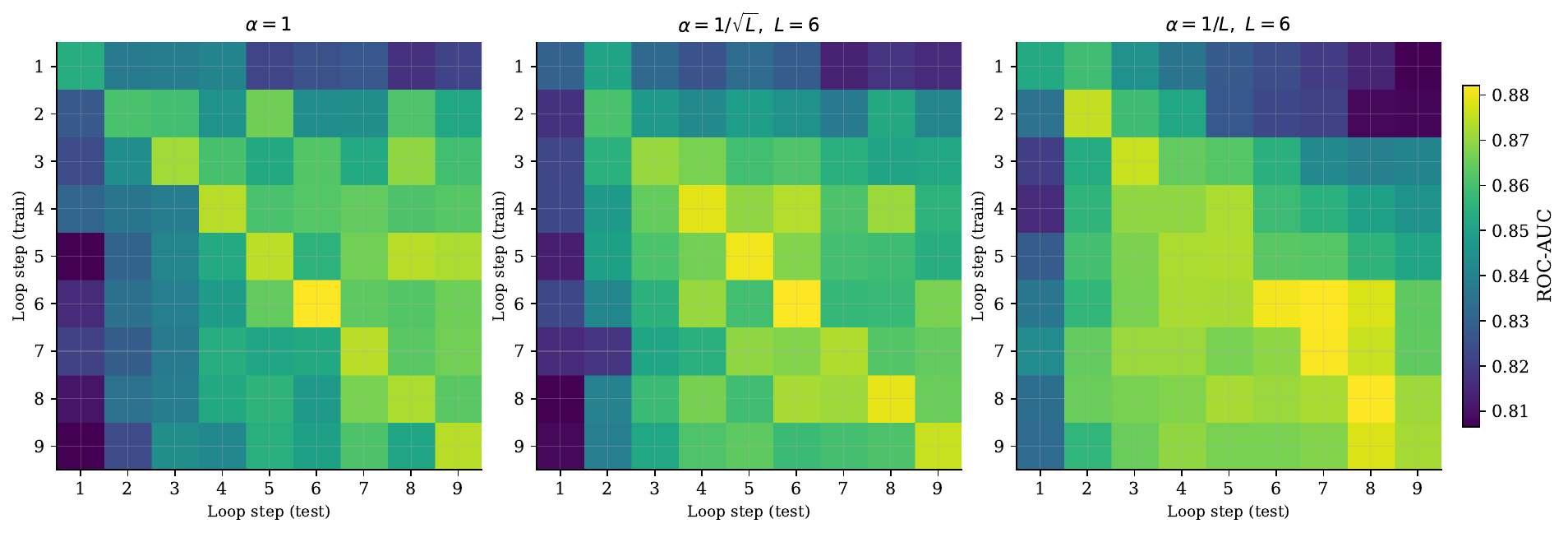}
\caption{Cross-step probing AUC for stacked blocks (\textit{left}) and looped models (\textit{right}).
Each entry $(i,j)$ shows the ROC-AUC of a logistic regression trained on embeddings at step $i$
and evaluated on embeddings at step $j$, averaged over all binary-classification datasets.}
\label{app:fig:embedding_probing}
\end{figure}

\boldheading{Probing classifiers.} As shown in Figure~\ref{app:fig:embedding_probing}, stacked blocks exhibit near-chance off-diagonal AUC, indicating limited transferability of representations across blocks. In contrast, looped models retain high off-diagonal AUC, showing that a classifier trained at one loop step generalizes well to the others. More interestingly, among looped models, the \emph{asymmetry} between the upper ($j > i$, train early/test late) and lower ($j < i$, train late/test early) triangles reveals distinct convergence regimes. For $\alpha = 1$, the upper triangle dominates: early-step features are largely preserved at later iterations, while later steps introduce additional discriminative structure that early classifiers cannot capture. For $\alpha = \tfrac{1}{L}$, this pattern reverses, becoming particularly pronounced beyond loop step $L$. Classifiers trained after step $L$ continue to transfer to earlier steps, whereas classifiers trained at earlier steps fail to transfer beyond step $L$. This asymmetry suggests that the scaled residual updates progressively overwrite features established during the first $L$ loops, providing a possible explanation for the degradation observed when models trained with a fixed $L$ are evaluated for more than $L$ iterations at test time.

\begin{figure}[htpb]
\centering
\includegraphics[height=0.2\linewidth]{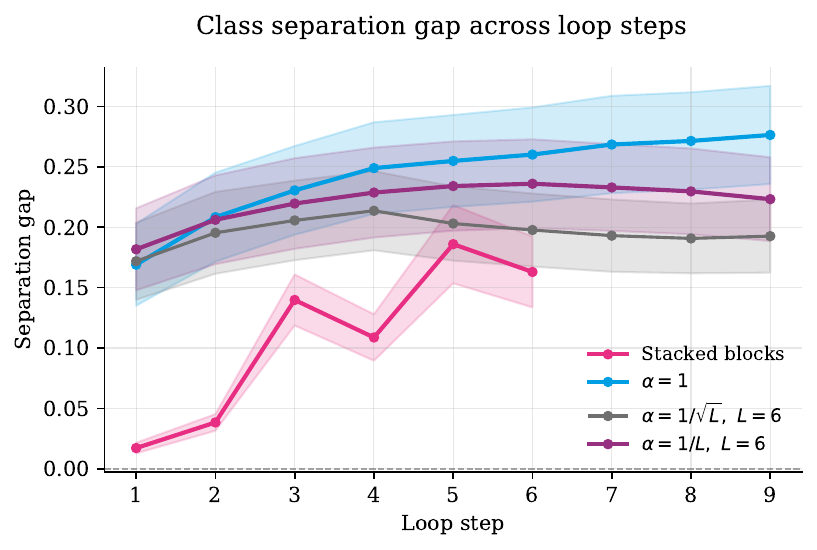}
\includegraphics[height=0.2\linewidth]{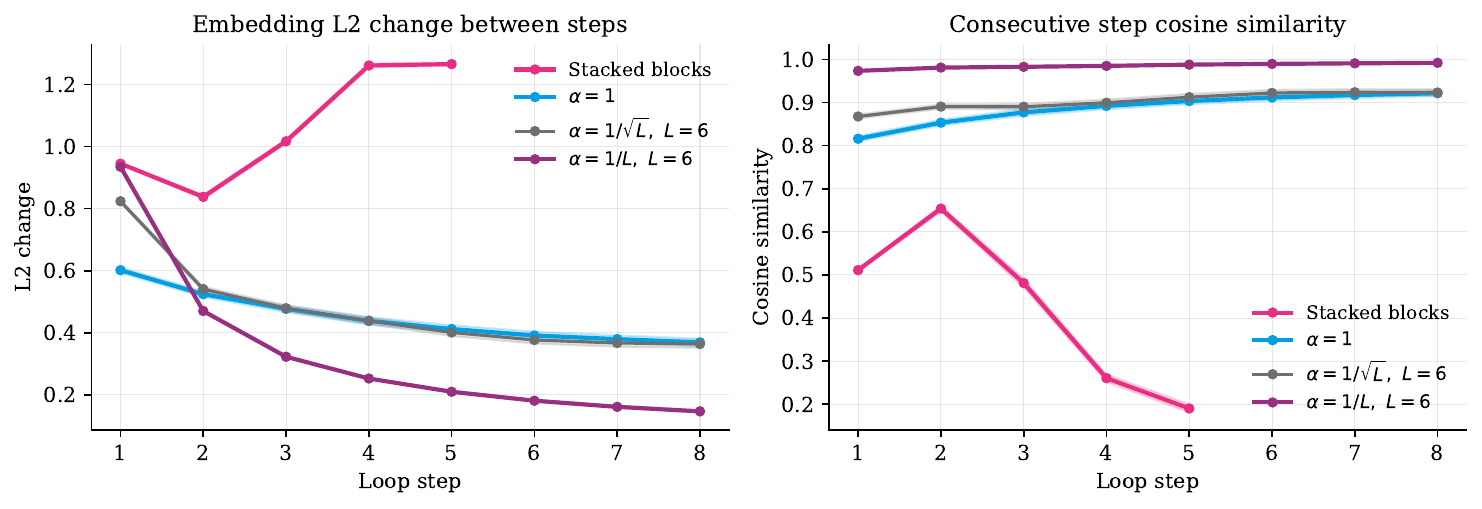}
\caption{
\textbf{Left}: class separation gap (mean inter-class minus intra-class cosine distance)
at each loop step. \textbf{Right}: embedding update magnitude across iterations,
measured as the normalised $\ell_2$ change between consecutive steps (\textit{left panel})
and as the cosine similarity between consecutive embeddings (\textit{right panel}).}
\label{app:fig:separation_gap}
\end{figure}

\boldheading{Separation gap.}
Figure~\ref{app:fig:separation_gap} shows that stacked blocks begin with a near-zero separation gap, indicating that the early layers primarily perform \emph{latent mapping} before progressively building class structure across subsequent layers. In contrast, looped models already exhibit a substantial separation gap at the first iteration, suggesting that the subsequent iterations primarily perform \emph{feature engineering}. For $\alpha=1$, the separation gap grows monotonically across all nine steps ($0.169 \to 0.276$). For $\alpha=\tfrac{1}{\sqrt{L}}$, the gap plateaus around step 4, while for $\alpha=\tfrac{1}{L}$ it peaks at step $L=6$ before marginally declining. The latter behavior is consistent with the representational overwriting observed in Figure~\ref{app:fig:embedding_probing}.

\begin{figure}[htpb]
\centering
\includegraphics[width=0.95\linewidth]{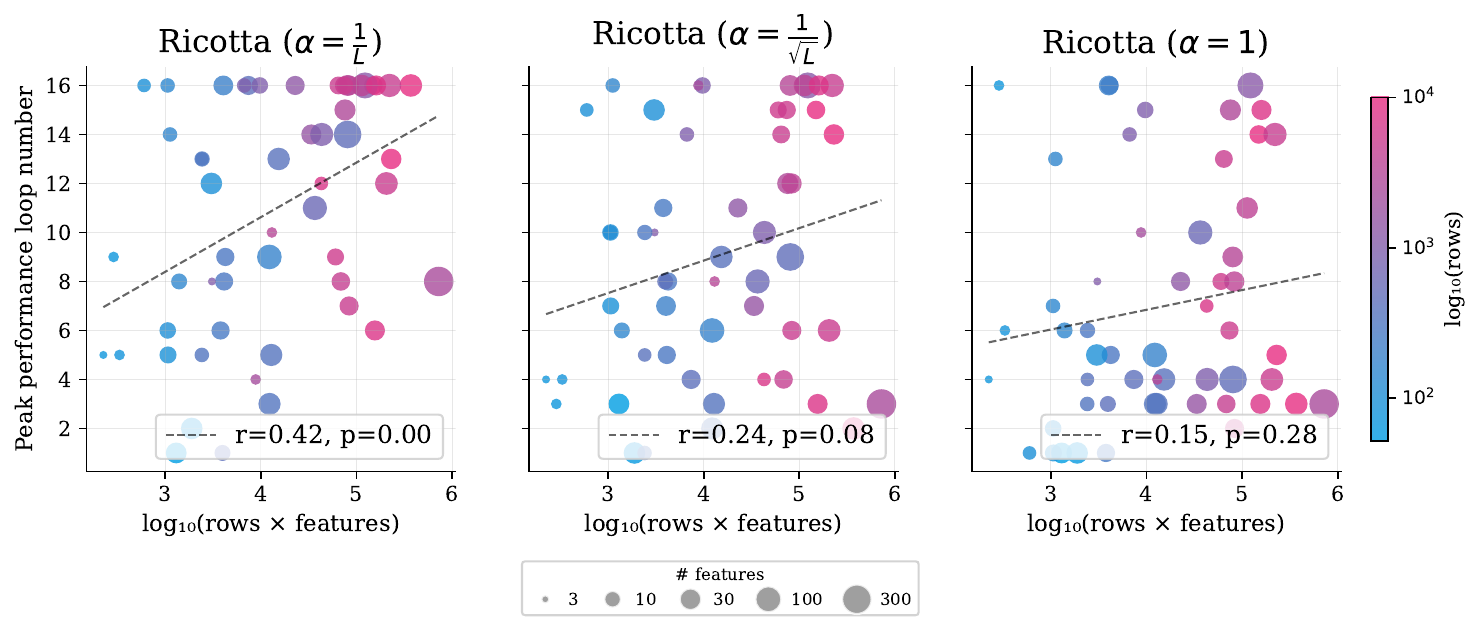}
\caption{With residual scaling $\frac{1}{L}$ we see a correlation between the optimal number of loops and dataset size.}
\label{app:fig:ablation_peak_loop}
\end{figure}

\boldheading{Residual Scaling.} Interestingly, as shown in Figure~\ref{app:fig:ablation_peak_loop}, with residual scaling $\alpha=1/L$, the optimal number of loops for a fixed residual depth ($L=16$) exhibits a clear correlation with dataset size. In contrast, we do not observe the same relationship without residual scaling or with $\alpha=1/\sqrt{L}$. This suggests that residual scaling by $1/L$ may help the model adapt its effective computation depth to the size of the dataset.

\subsection{PCA analysis}
\label{app:pca_analysis}

To characterize how information is organized within the iterative latent space, we apply PCA to the test-row embeddings at each loop step. For each dataset, embeddings are averaged across repeats within a fold and stacked into an $SN \times D$ matrix (loop steps $\times$ test samples), then a PCA is fit retaining the top 10 principal components.

We score each PC $k$ against four candidate roles:
\begin{itemize}

\item \textbf{Time-step ($t$):} absolute Pearson correlation between PC scores and the loop-step index,
\[
  A_k^{(t)} = \left|\rho(z_k,\, t)\right|.
\]

\item \textbf{Class separation ($y$):} eta-squared measuring how much variance in PC scores is explained by the binary class label,
\[
  A_k^{(y)} = \eta^2(z_k,\, y)
    = \frac{\displaystyle\sum_{c\in\{0,1\}} n_c\bigl(\bar{z}_{k,c} - \bar{z}_k\bigr)^2}
           {\displaystyle\sum_j \bigl(z_{k,j} - \bar{z}_k\bigr)^2},
\]
where $n_c$ and $\bar{z}_{k,c}$ are the count and mean PC score for class $c$, and $\bar{z}_k$ is the overall mean.

\item \textbf{Confidence ($H$):} absolute Pearson correlation between PC scores and per-sample prediction entropy,
\[
  A_k^{(H)} = \left|\rho(z_k,\, H)\right|, \qquad
  H_i = -\sum_{c=1}^{C} p_{ic}\log p_{ic},
\]
averaged across repeats and loop steps.

\item \textbf{Sample identity ($i$):} eta-squared measuring how much variance is explained by sample membership,
\[
  A_k^{(i)} = \eta^2(z_k,\, i)
    = \frac{S\displaystyle\sum_{r=0}^{N-1}\bigl(\bar{z}_{k,r} - \bar{z}_k\bigr)^2}
           {\displaystyle\sum_j \bigl(z_{k,j} - \bar{z}_k\bigr)^2},
\]
where $\bar{z}_{k,r}$ is the mean PC score for sample $r$ across its $S$ loop steps.

\end{itemize}

For each role, the 10 raw scores are L1-normalized to form a probability distribution $p_r \in \Delta^9$ over PC indices, indicating the relative affinity of each PC for that role. The four distributions are computed independently, so a single PC may carry signal for multiple roles.

\begin{figure}[htpb]
\centering
\includegraphics[width=0.95\linewidth]{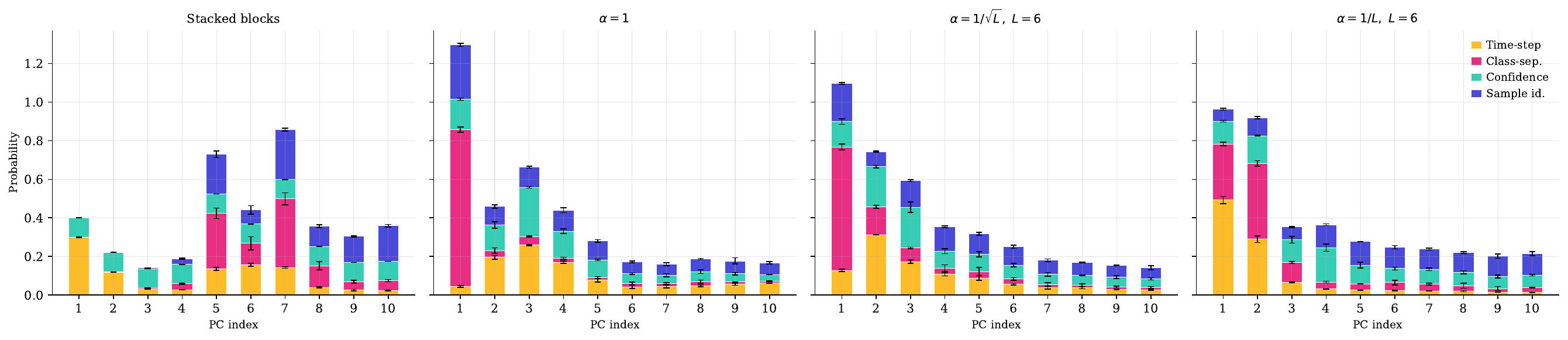}
\caption{\textbf{Structure of information in the iterative latent space.} We apply PCA to query embeddings pooled across loop steps and samples, and measure the affinity of each principal component (PC) with four functional roles: time-step progression, class separation, prediction confidence, and sample identity.}
\label{app:fig:pc_role_histogram}
\end{figure}

Within each cross-validation fold, the per-dataset probability vectors are averaged across datasets to yield one vector per role per fold. We report the mean and $95\%$ confidence interval ($\pm 1.96 \times \text{SEM}$) across the three folds (Figure ~\ref{app:fig:pc_role_histogram}).

As shown in Figure~\ref{app:fig:pc_role_histogram}, in recurrent model variants the time-step role is strongly concentrated in the leading PC, followed by class-separation signal in the next principal components — indicating that the model organizes its iterative dynamics along a small number of structured axes. Interestingly, in the stacked-blocks variant the contributions of all four roles are more diffuse, spreading across higher-order components rather than concentrating in the first few.

\subsection{Performance}
\label{app:performance}
\begin{figure}[htpb]
\centering
\includegraphics[width=0.95\linewidth]{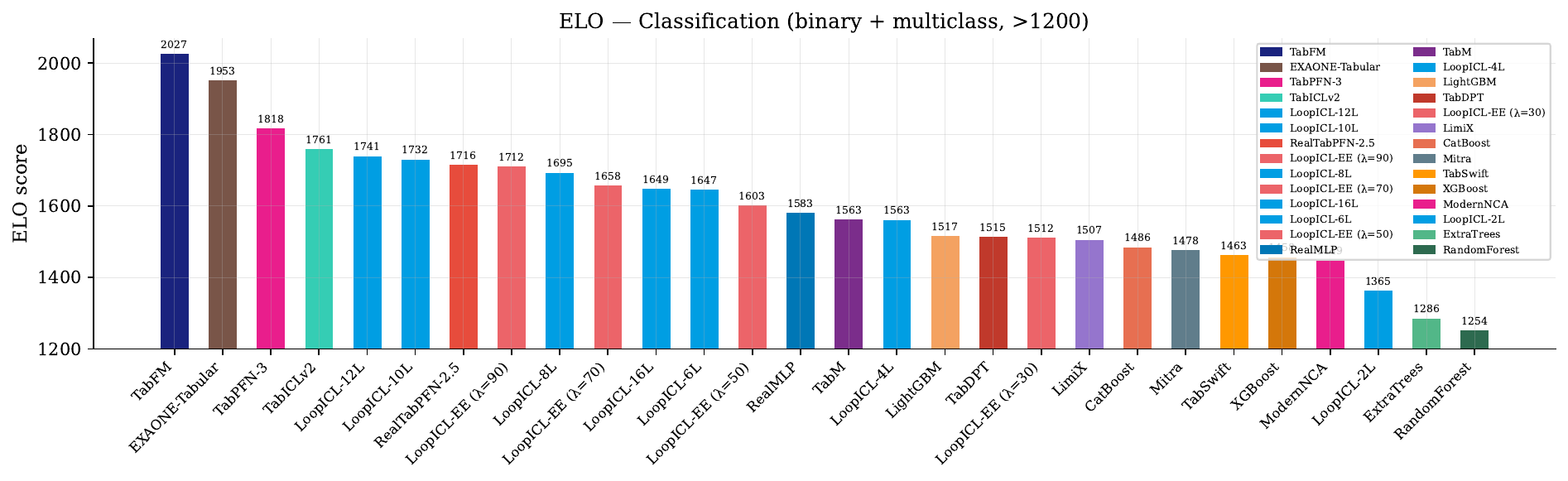}
\caption{Elo ratings for classification on \TabArena{}. Higher is better.}
\label{app:fig:tabarena_elo_classification}
\end{figure}

\begin{figure}[htpb]
\centering
\includegraphics[width=0.95\linewidth]{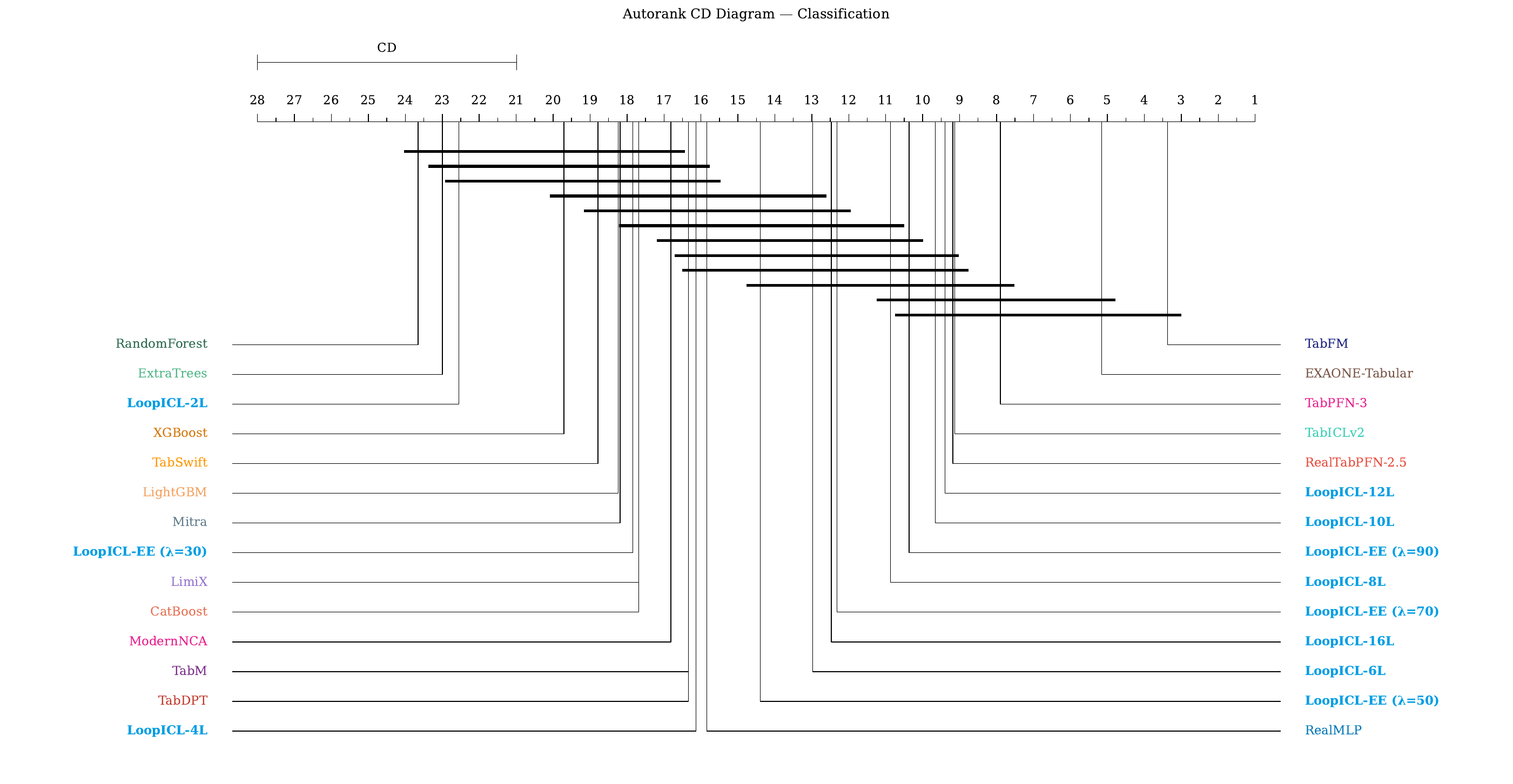}
\caption{Critical difference diagram for classification on \TabArena{}, computed via the \texttt{autorank} framework~\citep{herbold2020autorank} using a Wilcoxon signed-rank test with Holm correction ($\alpha=0.05$). Models connected by a horizontal bar are not significantly different.}
\label{app:fig:tabarena_autorank_classification}
\end{figure}

\begin{figure}[htpb]
\centering
\includegraphics[width=0.95\linewidth]{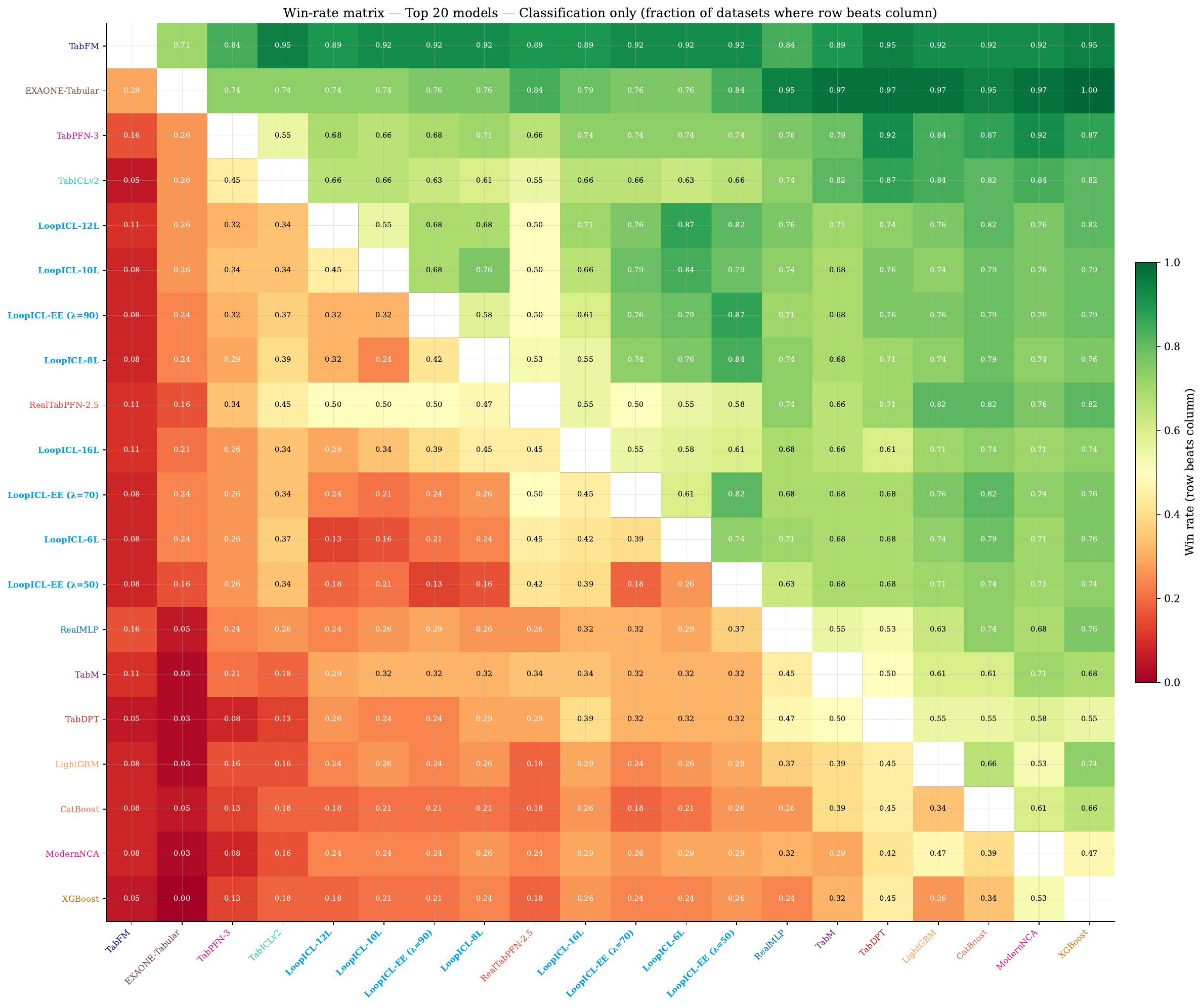}
\caption{Pairwise win rates on \TabArena{} classification (top-20 models). Each cell shows the fraction of datasets where the row model outperforms the column model.}
\label{app:fig:tabarena_winrate_top20}
\end{figure}

\begin{figure}[htpb]
\centering
\includegraphics[width=0.95\linewidth]{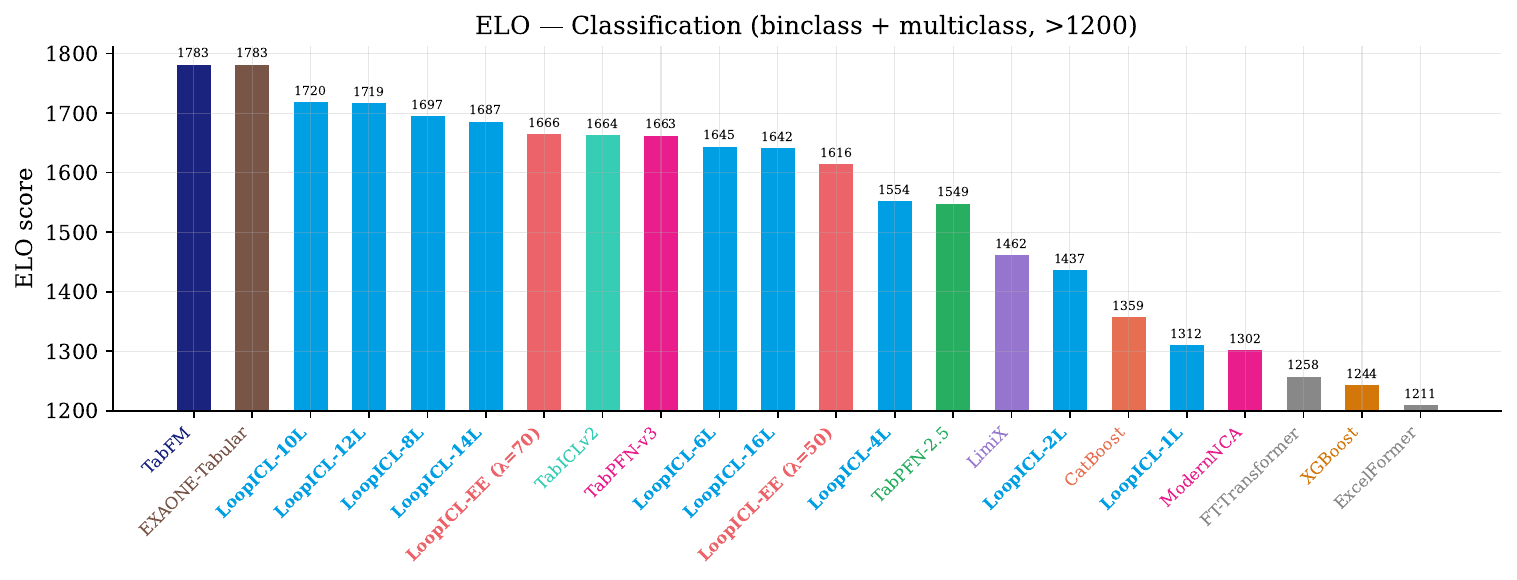}
\caption{Elo ratings for classification on \Talent{}. Higher is better.}
\label{app:fig:talent_elo_classification}
\end{figure}

\begin{figure}[htpb]
\centering
\includegraphics[width=0.95\linewidth]{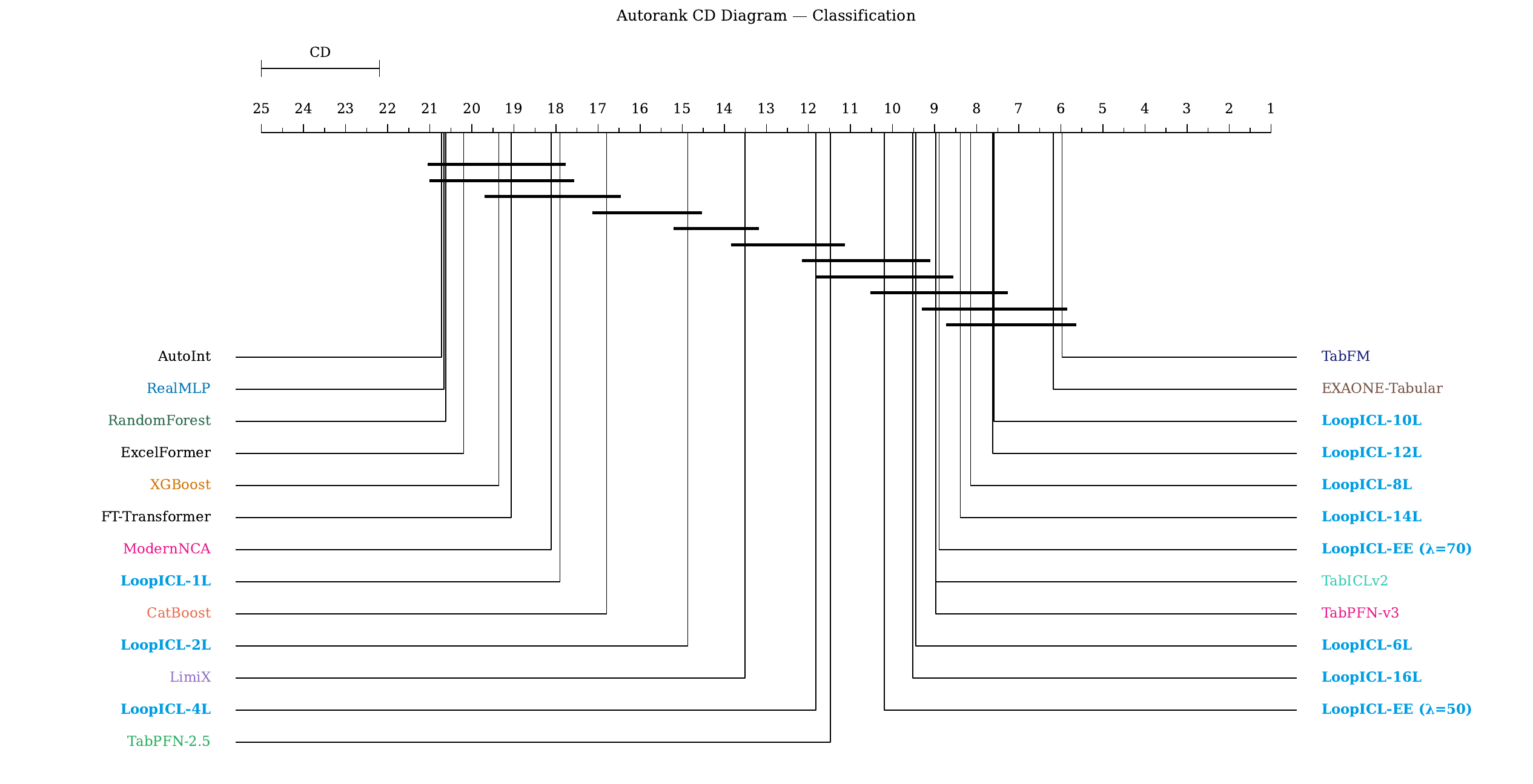}
\caption{Critical difference diagram for classification on \Talent{}, computed via the \texttt{autorank} framework~\citep{herbold2020autorank} using a Wilcoxon signed-rank test with Holm correction ($\alpha=0.05$). Models connected by a horizontal bar are not significantly different.}
\label{app:fig:talent_autorank_classification}
\end{figure}

\begin{figure}[htpb]
\centering
\includegraphics[width=0.95\linewidth]{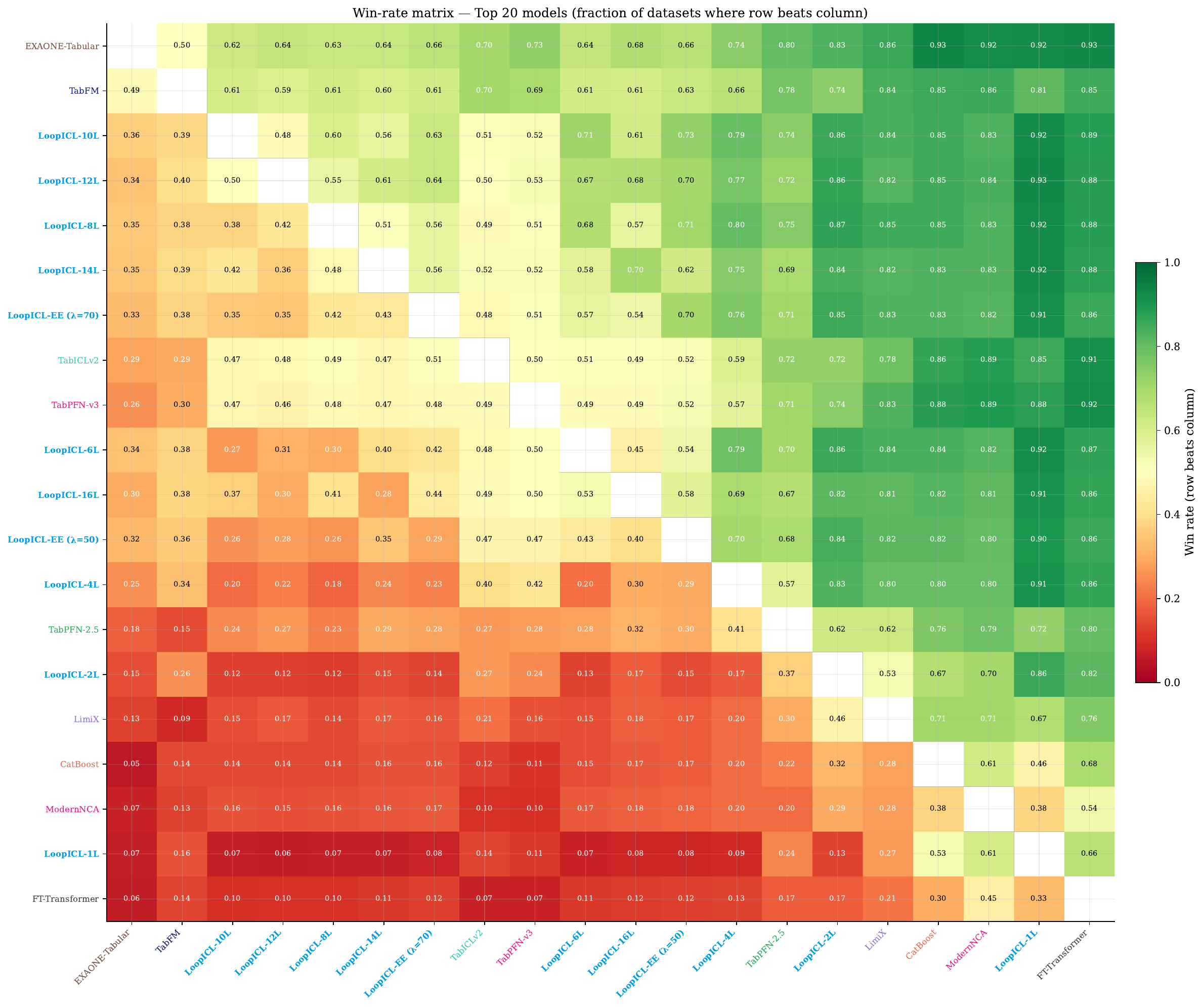}
\caption{Pairwise win rates on \Talent{} classification (top-20 models). Each cell shows the fraction of datasets where the row model outperforms the column model.}
\label{app:fig:talent_winrate_top20}
\end{figure}

Here we provide more details on the performance on \TabArena{} and \Talent{}.
For \TabArena{}, we report Elo ratings (Figure~\ref{app:fig:tabarena_elo_classification}), a critical difference diagram (Figure~\ref{app:fig:tabarena_autorank_classification}), and pairwise win rates among the top-20 models (Figure~\ref{app:fig:tabarena_winrate_top20}).
For \Talent{}, we report the corresponding Elo ratings (Figure~\ref{app:fig:talent_elo_classification}), critical difference diagram (Figure~\ref{app:fig:talent_autorank_classification}), and pairwise win rates (Figure~\ref{app:fig:talent_winrate_top20}).

\end{document}